\documentclass{article}
\usepackage{amsmath}
\usepackage{bbold}
\usepackage{amsthm}
\usepackage{xcolor}
\usepackage{relsize}
\usepackage{blindtext}
\usepackage[T1]{fontenc}
\usepackage[utf8]{inputenc}
\usepackage{graphicx}
\usepackage{authblk}
\usepackage{wrapfig}
\usepackage{lipsum}
\usepackage{subcaption}
\usepackage{amssymb,amsmath,amsthm}
\usepackage{graphicx}
\DeclareMathOperator{\Hom}{Hom}
\DeclareMathOperator*{\argmin}{arg\,min}
\DeclareMathOperator*{\argmax}{arg\max}
\usepackage{capt-of}
\usepackage{amssymb}
\newtheorem{theorem}{Theorem}
\usepackage{tikz-cd}
\usepackage{hyperref}
\usepackage{framed}
\graphicspath{ {./figures/} }
\usepackage[left=1.5cm, right=2cm]{geometry}

\title{ Multi-Irreducible Spectral Synchronization \\ for Robust Rotation Averaging }

\author{ Owen Howell$^1$$^3$, Haoen Huang$^2$,  and David Rosen$^1$$^2$ }
\date{%
	\small
	$^1$Northeastern University, Department of Electrical and Computer Engineering, 360 Huntington Ave, Boston, MA 02115, USA,\\%
	$^2$Northeastern University, Department of Mathematics, 360 Huntington Ave, Boston, MA 02115, USA,\\%
	$^3$University of Chicago, Department of Physics, 5801 S. Ellis Ave. Chicago, IL 60637\\%
	\today
}

\begin{document}
\maketitle
\begin{abstract}
Rotation averaging (RA) is a fundamental problem in robotics and computer vision.  In RA, the goal is to estimate a set of $N$ unknown orientations $R_{1},  ..., R_{N} \in SO(3)$, given noisy measurements $R_{ij} \sim R^{-1}_{i} R_{j}$ of a subset of their pairwise relative rotations. This problem is both nonconvex and NP-hard, and thus difficult to solve in the general case. We apply harmonic analysis on compact groups to derive a (convex) spectral relaxation constructed from truncated Fourier decompositions of the individual summands appearing in the RA objective; we then recover an estimate of the RA solution by computing a few extremal eigenpairs of this relaxation, and (approximately) solving a consensus problem.


Our approach affords several notable advantages versus prior RA methods: it can be used in conjunction with \emph{any} smooth loss function (including, but not limited to, robust M-estimators), does not require any initialization, and is implemented using only simple (and highly scalable) linear-algebraic computations and parallelizable optimizations over band-limited functions of individual rotational states. Moreover, under the (physically well-motivated) assumption of multiplicative Langevin measurement noise, we derive explicit performance guarantees for our spectral estimator (in the form of probabilistic tail bounds on the estimation error) that are parameterized in terms of graph-theoretic quantities of the underlying measurement network. By concretely linking estimator performance with properties of the underlying measurement graph, our results also indicate how to devise measurement networks that are \emph{guaranteed} to achieve accurate estimation, enabling such downstream tasks as sensor placement, network compression, and active sensing.

\end{abstract}    

\section{Introduction}


\emph{Rotation averaging} (RA) is a fundamental problem in robotics and computer vision, lying at the core of many geometric estimation tasks such as bundle adjustment \cite{Triggs2000Bundle}, structure from motion \cite{Tron2016Survey}, and simultaneous localization and mapping (SLAM) \cite{Cadena_2016}.  In particular, the ability to recover high-fidelity solutions of SLAM and RA problems is essential for functional mobile robots \cite{Rosen_2021,Cadena_2016,Placed_2022}.

Simultaneous Localization and Mapping (SLAM) is the process of jointly estimating the configuration of a robot and its environment, given a set of noisy sensor measurements. High-fidelity SLAM is an integral component of functional mobile robots \cite{Rosen_2021,Cadena_2016,Placed_2022}. The rotation averaging (RA) problem is a sub-problem of the general SLAM problem that is concerned with estimating a set of rotations given noisy samples of a subset of their pairwise relative differences.  The difficulty of the general SLAM problem is due to the difficulty of the RA subproblem specifically: indeed, in standard formulations of SLAM, it is possible to solve for the optimal values of the translational states as functions of the rotational estimates, thereby analytically eliminating the translations from the problem \cite{Rosen_2019}. 

However, rotation averaging is also computationally hard to solve in general \cite{Hartley_2013}.  Specifically, the RA problem is naturally formulated as a maximum likelihood estimation over the special orthogonal group $SO(d)$.  Because $SO(d)$ is a non-convex set, the RA estimation problem is also non-convex, with many local minima that can entrap local optimization methods.  Standard RA algorithms (based on local optimization) are therefore not guaranteed to recover  globally optimal solutions: instead, the solutions that they return depend on the initial choice of rotation estimates.

\begin{figure*}
\centering
\begin{subfigure}{1.0\linewidth}
\fbox{\includegraphics[width=1.0\textwidth]{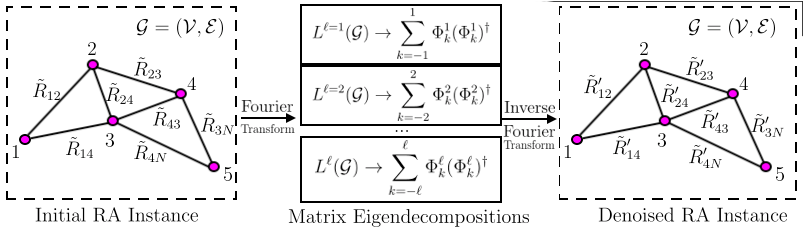}}
\end{subfigure}
\caption{ $SO(3)$ Multi-Irreducible Spectral Relaxation: An instance of an RA problem on graph $\mathcal{G} = ( \mathcal{V} , \mathcal{E} )$ with edges $\tilde{R}_{ij}$. The Fourier Transform on $SO(3)$ \ref{Fourier Transform on Groups} of the loss function is computed to construct a set of symmetric matrices $L^{\ell}(\mathcal{G})$, one for each $SO(3)$-irreducible $\ell = 1, 2, 3, ... $. We then compute the $2\ell+1$ lowest eigenvectors of the matrix $L^{\ell}(\mathcal{G})$. We then inverse Fourier transform to compute a new set of denoised edge rotations $\tilde{R}'_{ij}$. Any rotation averaging method can then be used to estimate the rotations from the denoised edge rotations $\tilde{R}'_{ij}$. When the edge loss is quadratic, only $L^{1}(\mathcal{G})$ is non-zero, and our method reduces to \cite{Doherty_2022}. The Multi-Irreducible Spectral Relaxation can be used for \emph{any} loss function including, but not limited to,
robust $M$-estimators .  }
\end{figure*}

Consequently there has been considerable interest in developing heuristic and approximate methods for finding high-quality initial rotation estimates.  In particular, the recent work \cite{Doherty_2022} proposed a simple initialization scheme based upon a \emph{spectral relaxation} of the RA estimation problem.  This approach only requires calculating the $d$ algebraically-smallest eigenvectors of a symmetric positive-definite matrix, which can be achieved using any matrix-power method, such as the Lanczos method \cite{Boumal_2013,Shuyang_2020,Belkin_2017}.  Furthermore, this method admits \emph{explicit}, \emph{computable}, \emph{non-asymptotic bounds} on the estimation error that are parameterized in terms of spectral graph-theoretic quantities of the underyling measurement graph.  Unfortunately, as was observed in \cite{Doherty_2022}, these bounds turn out to be quite loose (i.e.\ extremely pessimistic) in practice. In this work, we propose a group-theoretic generalization of the spectral initialization \cite{Doherty_2022} that we call \emph{multi-irreducible spectral synchronization}.  Specifically, we apply harmonic analysis on compact groups to derive a (convex) spectral relaxation constructed from truncated Fourier decompositions of the individual summands appearing in the RA objective.  

Our approach improves upon prior work in rotation averaging along several important axes:



\begin{itemize} 
\item Our approach generalizes \cite{Doherty_2022} to admit the use of \emph{any} smooth loss function (including robust M-estimators).\footnote{Our method reduces to \cite{Doherty_2022} when using a squared Frobenius norm loss.}
\item Our method provides a simple mechanism for trading off estimation accuracy with computational cost, by varying the number of irreducibles used in the truncated Fourier decomposition.  In particular, we show that using multiple irreducible $SO(3)$ representations for rotation averaging significantly enhances estimation accuracy, especially in the high-noise regime.
\item Under the  (physically well-motivated) assumption of multiplicative Langevin noise, we derive sharp computable \emph{a priori} performance guarantees that are parameterized in terms of spectral graph-theoretic quantities of the underlying measurement network. In particular, our results tighten the bounds reported in \cite{Doherty_2022} by several orders of magnitude.

\item We demonstrate that our method achieves state of the art performance on both real and synthetic rotation averaging problems constructed from the large-scale benchmarks \cite{Carlone_2015_Lagrangian}, significantly outperforming prior techniques with large improvement in the outlier contaminated regime.


\end{itemize}


\section{Related Work}

\subsection*{Rotation averaging}

The most common approach to rotation averaging is to formulate the problem as a maximum likelihood estimation (or more generally M-estimation), and then apply local optimization methods (such as gradient descent or quasi-Newton methods) to recover a \emph{local} solution of the estimation problem \cite{Hartley_2013}.  This approach is attractive from a computational standpoint, as local optimization methods can scale gracefully to even very large problem sizes \cite{Agarwal_2011_Rome}.  However, this comes at the expense of \emph{reliability}: because the RA problem is nonconvex, this approach can only guarantee the recovery of a \emph{local} (rather than \emph{global}) solution.  The estimate that these methods ultimately return depends upon the initial point at which the local search was started.

To address this pitfall, several lines of recent work have addressed the convergence properties of optimization methods used in rotation averaging.  One line develops inexpensive heuristic approximation schemes that can produce a (hopefully) high-quality approximate solution, which can be taken as an estimate in its own right or used to initialize a subsequent nonlinear refinement \cite{Carlone2015Initialization,Martinec2007Robust,Doherty_2022,Moreira2021fast,Singer2011Angular,Arrigoni2020Synchronization}.  While these often work remarkably well in practice, in general they too lack any kind of performance guarantees.

Another recent line of work has explored the use of \emph{convex} (typically \emph{semidefinite}) \emph{relaxations} for recovering high-quality RA solutions \cite{Rosen_2019,eriksson2019rotation,Dellaert_2020_Shonan,Rosen_2021_Advances,Eriksson2018Rotation,Carlone_2015_Lagrangian}.  Remarkably, several recent works have shown that this approach enables the recovery of \emph{exact, globally optimal} solutions of the RA problem when the noise on the data is sufficiently small \cite{Rosen2016Certifiably,eriksson2019rotation}.  However, the optimality of the estimates that these methods produce can only be ascertained \emph{post hoc}, on a per-instance basis.  Moreover, the relaxations used in these specific techniques are closely tied to the use of the least-squares loss function, and so they are not robust to the presence of outlier contamination.  While it is (in principle) possible to extend these convex relaxation-based approaches to use more general classes of robust loss functions, at present the additional computational cost of solving these (more complex) relaxations is far beyond the capabilities of current off-the-shelf convex optimization tools \cite{Rosen_2021_Advances}.

Finally, in large-scale applications of rotation averaging it is often necessary to use robust loss functions to guard against possible outlier contamination in the measurements \cite{Triggs2000Bundle,Hartley2004Multiple, Huber2004Robust}.  Without this safeguard, even a small fraction of contaminated measurements can significantly degrade the resulting estimate.  Unfortunately, having favorable outlier attenuation properties (i.e.\ finite rejection) requires the use of robust loss functions that are significantly nonconvex, making the underlying RA M-estimation even more challenging to solve with local optimization; this is further exacerbated by the fact the the initialization strategies typically used in combination with local optimization are also not robust to outlier contamination.

\begin{wrapfigure}{r}{0.5\textwidth}
	\centering
	\fbox{\includegraphics[width=0.5\textwidth]{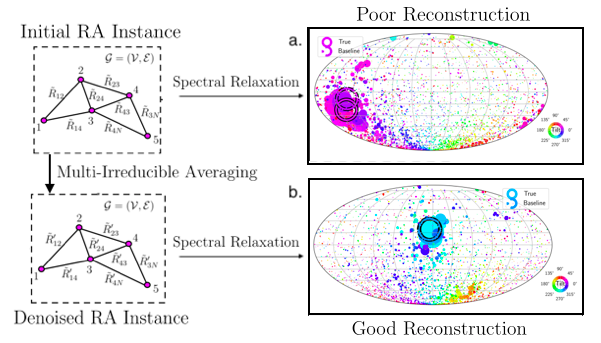}}
	\caption{ Under large noise, the naive spectral relaxation \cite{Doherty_2022} results in poor reconstruction. When the initial RA problem instance is averaged over multiple irreducible representations, the spectral relaxation produces high fidelity reconstructions, even in the high noise regime. Panel 'a' shows the spectral relaxation \cite{Doherty_2022} predicted and true rotations (dotted and bold annulus, respectively) as well as the distribution of predicted rotations. Panel 'b' shows the spectral relaxation \cite{Doherty_2022} predicted and true rotations (dotted and bold annulus, respectively) as well as the distribution of predicted rotations after the multi-irreducible averaging is performed. Note that after de-noising, the predicted and true rotations are identical. For more information about representing rotations on 2d plots, please see \cite{Murphy_2022_implicitpdf}.   }
	\vspace{-1.0cm}
\end{wrapfigure}

Our multi-frequency spectral estimator can be viewed as a convex relaxation of the RA synchronization problem.  However, in contrast to prior convex relaxation approaches for SLAM and RA (which are typically built using the algebraic structure of the problem, via moment or Lagrangian relaxation \cite{Rosen_2021_Advances}), our relaxation strategy is based on a truncated Fourier decomposition.  This enables us to seamlessly adapt our approach to a wide range of loss functions (including robust M-estimators).  Moreover, because our approach admits the use of robust estimators, it continues to provide useful estimates even in the presence of outlier contamination.

\subsection*{Cryogenic Electron Microscopy}

Cryogenic electron microscopy (cryo-EM) is a revolutionary technique for imaging biological macromolecules \cite{Dubochet_1981_Vitrification}. A key step in cryo-EM reconstruction is the co-registration of 2D projections of molecules imaged in initially unknown orientations, a problem which can be formulated as a 2D analogue (phase synchronization) of (3D) rotation averaging. Images produced by cryo-EM imaging have large amounts of noise, necessitating robust statistical estimation methods.

Our multi-frequency spectral synchronization approach is inspired by prior work in cryo-EM that likewise exploits group-theoretic structure \cite{Fan_2021_Representation}.  Methods like steerable PCA \cite{Zhao_2013_Fourier} utilize the $SO(2)$-symmetry inherent in cryo-EM imagining to learn de-noised images. \cite{Wang_2012} used vector diffusion maps \cite{Singer_2011_Vector} to model cryo-EM images as data points living in an $S^{2}$ latent space.  \cite{Hadani_2011_Representation} shows that encoding multiple non-equivalent representations improves the quality of de-noising. \cite{Bandeira_2015_Nonunique} used the Fourier decomposition on groups to construct an SDP. \cite{Gao_2019_Multifrequency} showed that de-noising methods that add multiple copies of nonequivalent irreducible are significantly more noise robust.

\section{Problem Formulation}

\subsection*{Lie Groups, Matrix Manifolds and Irreducible Representations}
The fundamental representation of  the rotation group in $d$-dimensions is given by 
\begin{align*}
	SO(d) = \{ \enspace R  \enspace | \enspace R\in \mathbb{R}^{d\times d} , \enspace R^{T}R = \mathbb{I}_{d} , \enspace \text{det}(R) = 1 \enspace \}.
\end{align*}
We will define the $N$-fold vectorization of the fundamental representation of $SO(d)$ as the manifold
\begin{align*}
SO(d)^{N} = \{ \enspace R \enspace | \enspace R = \begin{bmatrix}
R_{1},
R_{2},
... ,
R_{N}
\end{bmatrix} \text{ with } R_{i} \in SO(d) \enspace  \}
\end{align*}
which consists of $N$ row concatenations of special orthogonal matrices. It should be noted that a group is an abstract mathematical object and that the standard parameterization is a \emph{choice}. There are multiple non-equivalent representations. We also define higher order representations of both $SO(d)$. Specifically, let $\rho$ be a representation of $SO(d)$. Then, we may define
\begin{align*}
SO( d )^{N}_{\rho} = \{ \enspace \rho(g) \enspace  | \enspace  \rho(g) = [ \rho(g_{1}) , \rho(g_{2}) , ... , \rho(g_{N}) ] \enspace \}
\end{align*}
where each $g_{i} \in SO(d)$ is an abstract group element.

\paragraph{Stiefel Manifold}
For $d\leq N$, the Stiefel Manifold $St(d,N)$ is defined as the set of orthonormal $d$-frames in $\mathbb{R}^{N}$. $St(d,N)$ has a natural embedding as a sub-manifold of $\mathbb{R}^{d \times N}$.
\begin{align}\label{Stiefel Manifold Defintion}
	St( d , N ) = \{ \enspace V \enspace | \enspace V\in \mathbb{R}^{d \times N} ,  \enspace VV^{T} = \mathbb{I}_{d} \enspace \}
\end{align}
$St(d, N )$ can also be viewed as the homogeneous space resulting from the quotient of orthogonal groups $St(d, N) \cong O(N)/O(N-d)$. A vectorization of any representation of rotation matrices $\rho(g) = \begin{bmatrix}
	\rho( g_{1}),
	\rho( g_{2}),
	... ,
	\rho( g_{N})
\end{bmatrix}  \in SO(d)^{N}_{\rho}$ satisfies the relation
\begin{align}
	\nonumber	\rho(g)\rho(g)^{T} =  \sum_{ i = 1}^{N} \rho(g_{i})\rho(g_{i})^{T} = N \mathbb{I}_{d_{\rho}}.
\end{align}

Thus, every $ \rho(g) \in SO(d)^{N}_{\rho}$ is a scalar multiple of an element of $St(d_{\rho}, d_{\rho}N)$ with proportionality constant $\frac{1}{\sqrt{N}}$. Specifically,
\begin{align*}
	\text{ if } \rho(g) \in SO(d)^{N}_{\rho} \implies \frac{1}{\sqrt{N}} \rho(g) \in St(d_{\rho}, d_{\rho}N)
\end{align*}
This relation allows for a natural relaxation method which is a generalization of the convex relaxation suggested in \cite{Doherty_2022}.

\subsection*{The rotation averaging problem}

Let $R_{1} , R_{2} , ... R_{N} \in SO(d) $ be rotation matrices. In the Rotation Averaging (RA) problem, we are concerned with recovering each of the $R_{i}$ given noisy samples of some subset of the pairwise $R_{i}^{T} R_{j}$.  Let us write the true $R_{i}^{T} R_{j}$ group elements as
\begin{align}\label{Pairwise Rotation Differences}
	& R_{ij} \triangleq R_{i}^{T} R_{j}
\end{align}

It should be noted that left multiplication of each $R_{i}$ by an element of $SO(d)$ describes the same pairwise rotation differences. Let $O \in SO(d)$. Then, under the transformation $R_{i} \rightarrow O R_{i} $,
the pairwise differences, $R_{ij}$ \eqref{Pairwise Rotation Differences}, remain unchanged as 
\begin{align}\label{Gauge Invarience}
(OR_{i})^{T} (OR_{j} ) =  (R_{i})^{T} O^{T}OR_{j}  = (R_{i})^{T} R_{j} = R_{ij}.
\end{align}
Thus, solutions to the rotation averaging problem are only defined up to left multiplication by an element of $SO(d)$. 

We will denote the noisy measurements as $\tilde{R}_{ij}$. We assume a noise model given by
\begin{align}\label{Pairwise Differences Noisy}
	& \tilde{R}_{ij} = R_{ij} R^{\epsilon}_{ij}
\end{align}
The random variables $R^{\epsilon}_{ij}$ are group valued and are constrained to satisfy $R^{\epsilon}_{ij} = (R^{\epsilon}_{ji})^{T}$ but are otherwise drawn independently. Note that we will assume that $R_{ii} = \mathbb{I}_{d}$ without loss of generality.

\section{Loss Function}
Let $ \mathcal{G} = ( \mathcal{E} , \mathcal{V} )$ be a graph. The most general possible loss function $\mathcal{L}$ that depends only on the ratio $g^{-1}_{i}g_{j}$ defined over group $G$ on graph edges can be written as
\begin{align*}
\mathcal{L} =	\sum_{ ij \in \mathcal{E} } f_{ij}( g_{i}^{-1}g_{j} ; \tilde{g}_{ij} ) 
\end{align*}
with corresponding MLE optimization problem
\begin{equation*}
    \label{RA_problem_formulation}
    \min_{g_i \in G} \sum_{(i,j) \in E} f_{ij}(g_i^{-1} g_j; \tilde{g}_{ij}) \tag{RA}
\end{equation*}
where each $f_{ij}( \enspace \cdot \enspace  ; \tilde{g}_{ij}) \colon G \to \mathbb{R}$ is a real-valued function (parameterized by $\tilde{g}_{ij} \in G$) that penalizes the discrepancy between the \emph{predicted} relative transformation $g_{ij} \triangleq g_i^{-1} g_j \in G$ and the \emph{measured} relative transformation $\tilde{g}_{ij} \in G$. 

In practice, the edge loss $f_{ij}$ is often taken to be the negative log-likelihood for the noisy measurement $\tilde{g}_{ij} \in G$ associated with the $(i,j)$-edge in the measurement network under an assumed generative model \cite{Doherty_2022}, 
\begin{equation}
\label{negative_log_likelihood_loss}
f_{ij}(g_{ij}; \tilde{g}_{ij}) = -\log p(\tilde{g}_{ij} \mid g_{ij}),
\end{equation}
in which case \eqref{RA_problem_formulation} is a standard maximum-likelihood estimation. However, our formulation \eqref{RA_problem_formulation} also admits the use of more general M-estimators, including the use of \emph{robust} loss functions that need not arise as negative log-likelihoods of a generative model \eqref{negative_log_likelihood_loss}.

\subsection{Fourier Transform on Groups Decomposition}

We now apply the representation theory of compact groups to rewrite the objective of the rotation averaging problem \eqref{RA_problem_formulation} as a \emph{linear} function of irreducible representations. This is a generalization of the standard Fourier transform from $S^1 \cong SO(2)$ to general compact groups $G$ \ref{Fourier Transform on Groups}. Using the Peter-Weyl theorem \cite{Zee_2016_Group}, we can expand \emph{any} function $f: G \rightarrow \mathbb{C}$ in terms of $G$-irreducibles as
\begin{align*}
	f(g) = \sum_{ \rho \in \hat{G} } d_{\rho} \sum_{kk'=0}^{d_{\rho}} \hat{f}^{\rho}_{kk'}  \rho_{kk'}(g) 
\end{align*}
where for each irreducible $\rho$ the $\hat{f}^{\rho}$ are $d_{\rho} \times d_{\rho}$ matrices. The expansion matrices $\hat{f}^{\rho}$ are given by
\begin{align*}
	\hat{f}^{\rho}_{kk'} = \int_{g \in G}dg \text{ } f(g) \rho_{kk'}(g^{-1})
\end{align*}
where $g$ is the Haar measure on $G$. Using this expansion, we have that,
\begin{align*}
\mathcal{L} = \sum_{ ij \in \mathcal{E} } f_{ij}( g_{i}g^{-1}_{j} ) =   \sum_{ ij \in \mathcal{E} } \sum_{ \rho \in \hat{G} } d_{\rho}\text{Tr}[  \hat{f}^{\rho}_{ij}\rho( g_{i}g^{-1}_{j} ) ] 
\end{align*}
We can further simplify this expression using the definition of a group representation $ \rho(gg') = \rho(g) \rho(g')$,
\begin{align*}
\mathcal{L} = \sum_{ ij \in \mathcal{E} } f_{ij}( g_{i}g^{-1}_{j} )  = \sum_{ ij \in \mathcal{E} } \sum_{ \rho \in \hat{G} } d_{\rho} \text{Tr}[  \hat{f}^{\rho}_{ij} \rho( g_{i} ) \rho(g^{-1}_{j} ) ]
\end{align*}
Note that for each $G$-irreducible $\rho$ each $\hat{f}^{\rho}_{ij}$ implicitly depend on the observed $\tilde{g}_{ij}$. Finally, re-indexing this sum, we may rewrite the \ref{RA_problem_formulation} loss function as 
\begin{align*}
\mathcal{L}  =  \sum_{ \rho \in \hat{G} } d_{\rho} \text{Tr}[  \sum_{ ij \in \mathcal{E} }   \rho( g_{i} )  \hat{f}^{\rho}_{ij}  \rho(g^{-1}_{j} ) ] 
\end{align*}
This maximum likelihood estimation problem can be rewritten as a group valued minimization problem, subject to the constraints that $\rho(g)$ is a valid element of the $\rho$ representation of $G$. Specifically, the optimal group elements can be written as an argmax over the group $G$,
\begin{align*}
\hat{g} = \argmax_{ g \in G^{N} } [ \enspace \sum_{ \rho \in \hat{G} } d_{\rho} \text{Tr}[  \sum_{ ij \in \mathcal{E} }   \rho( g_{i} )  \hat{f}^{\rho}_{ij}  \rho(g^{-1}_{j} ) ] \enspace  ]
\end{align*}

\subsubsection{Independence Irreducible Relaxation }
We will first relax each $\rho(g)$ to be \emph{independent} group elements. For each irreducible $\rho$, we then have a independent maximization problem which returns a value
\begin{align*}
	\hat{g}^{\rho} = \argmax_{ g \in G^{N} }  \text{Tr}[  \sum_{ ij \in \mathcal{E} }   \rho( g_{i} )  \hat{f}^{\rho}_{ij}  \rho(g^{-1}_{j} ) ]  
\end{align*}
Note that this problem still involves a optimization over $G^{N}$, and is in general intractable. However, we can solve each irreducible problem using the method proposed in \cite{Doherty_2022}. Let us vectorize each representation $\rho(g) = [ \rho(g_{1}) , \rho(g_{2}) , ... , \rho(g_{N} ) ]$. For all $g\in G$, each representation $\rho(g)$ is an unitary matrix. The vectorization $\rho(g)$ satisfies
\begin{align*}
	\rho(g) \rho(g)^{T} = N \cdot \mathbb{I}_{d_{\rho}}
\end{align*}
Using this notation, each $\rho$ sub-problem can written as,
\begin{align*}
\hat{g}^{\rho} = \argmin_{g \in G^{N}} \text{Tr}[ L^{\rho}( \mathcal{G} ) \rho(g)^{T}\rho(g)  ]
\end{align*}
where, following the nomenclature of \cite{Doherty_2022,Wang_2012}, $L^{\rho}( \mathcal{G})$ is termed the $\rho$-Laplacian. Each $L^{\rho}( \mathcal{G} )$ is a symmetric $d_{\rho} N \times d_{\rho} N$ matrix with $d_{\rho} \times d_{\rho}$ blocks given by $L^{\rho}( \mathcal{G} )_{ij} = \hat{f}^{\rho}_{ij}$. Note that $\hat{f}^{\rho}_{ij}$ implicitly depend on the observed $\tilde{g}_{ij}$. For the class of loss functions that we are interested in, the form of $\hat{f}^{\rho}_{ij}$ can be further simplified. The loss functions $f$ that we will be interested in take the form
\begin{align*}
f_{ij}(g_{i},g_{j} ; \tilde{g}_{ij} ) = h_{ij}( || \sigma(g^{-1}_{i}g_{j}) - \sigma( \tilde{g}_{ij}) || )
\end{align*}
where $\sigma$ is a representation of $G$, $|| \cdot ||$ is a unitary norm and $h_{ij}: \mathbb{R}^{+} \rightarrow \mathbb{R}^{+} $ is a positive scalar function.

\begin{theorem}\label{Main:Theorem_I}
Let $h : \mathbb{R}^{+} \rightarrow \mathbb{R}^{+}$ be a smooth function. Let $(\sigma , V)$ be any representation of the group $G$. Let $|| \cdot ||$ be a unitary norm on the vector space $V$. Let $f$ be a group valued function defined as
\begin{align*}
f(g_{a},g_{b} ;\tilde{g}_{ab} ) = h(  || \sigma(g^{-1}_{a}g_{b}) - \sigma(\tilde{g}_{ab} ) ||   )
\end{align*}
Then, the Fourier Transform of $f$ can be written as $ \hat{f}^{\rho} = K^{\rho}\rho( \tilde{g}_{ab} )$, where $K^{\rho} \in \mathbb{C}$ is a constant given by
\begin{align*}
K^{\rho} = \frac{1}{d_{\rho}} \int_{g\in G}dg \text{ } h( || \sigma(g) - \mathbb{I}_{d_{\sigma}} || ) \chi^{\rho}(g) 
\end{align*}
where $\chi^{\rho}(g) = \text{Tr}[ \rho(g) ]$ is the character of the $\rho$ representation. The integral expression for $K^{\rho}$ is invariant under the normal transformation $g \rightarrow hgh^{-1}$. If the group $G$ is continuous, the expression for $K^{\rho}$ and can be simplified using the Weyl integration formula.
\end{theorem}

The proof of theorem \ref{Main:Theorem_I} is given in \ref{Suppl_Thrm_I}. Theorem \ref{Main:Theorem_I} gives a description of $L^{\rho}(\mathcal{G})$. Let 
\begin{align*}
K^{\rho}_{ij} = \frac{1}{d_{\rho}} \int_{g\in G}dg \text{ } h_{ij}( || \sigma(g) - \mathbb{I}_{d_{\sigma}} || ) \chi^{\rho}(g) 
\end{align*}
and define the $i$-th node $\rho$-degree as 
\begin{align*}
D^{\rho}_{i}  = \sum_{ j\ne i } k^{\rho}_{ij}
\end{align*}
Then, each $\rho$-Laplacian has $d_{\rho} \times d_{\rho}$ blocks given by
\begin{align}\label{Definition of Noisy Laplcian Matrix}
L^{\rho}( \mathcal{G} )_{ij} = \begin{cases}
&  D^{\rho}_{i} \mathbb{I}_{ d_{\rho} } \text{ if } i=j \\ 
& K^{\rho}_{ij} \rho( \tilde{g}_{ij} ) \text{ if } i \ne j
\end{cases}
\end{align}
For loss functions $h_{ij}$ that we are interested in, the $K^{\rho}_{ij}$ are positive definite. Using Cesàro summation, there is always a way to re-sum terms to get positive definite $K^{\rho}_{ij}$ coefficients \cite{Bandeira_2015_Nonunique,Gao_2019_Multifrequency} (see \ref{Cesàro Summation and Fejér Kernels}).

\section{Multi-Irreducible Spectral Relaxation}

Note that for each irreducible representation $\rho$, the matrix $\rho(g)$ satisfies $\rho(g)\rho(g)^{T} = N \cdot \mathbb{I}_{d_{\rho}}$. Thus, $\frac{1}{\sqrt{N}} \rho(g)$ is an element of $St(d_{\rho},d_{\rho}N)$. We can relax the $\rho(g) \in \rho(G)^{N}$ constraint in the MLE problem to the weaker constraint that $\frac{1}{\sqrt{N}} \rho(g) \in St(d_{\rho},d_{\rho}N)$. Consider the set of less restrictive optimization problems
\begin{align}\label{Relaxed MLE}
	& \Phi^{\rho} = \min_{ \rho(g) \in \mathbb{R}^{d_{\rho}\times d_{\rho}N} } \text{Tr}[ L^{\rho}(\mathcal{G}) \rho(g)^{T}\rho(g) ] \\
	& \nonumber  \text{s.t. } \rho(g)\rho(g)^{T}  = N \cdot \mathbb{I}_{d_{\rho}} 
\end{align} where the $\rho(g) \in \rho(G)^{N}$ constraint has been replaced by the condition $\rho(g)\rho(g)^{T} = N \cdot \mathbb{I}_{d_{\rho}} $. This relaxed problem is a matrix eigenvalue problem. 

\begin{theorem}
Let $(\rho , V)$ be an irreducible representation of the group $G$. Then, the graph Laplacian $L^{\rho}( \mathcal{G} )$ is a $d_{\rho}N \times d_{\rho}N$ Hermitian matrix with a $d_{\rho}$-fold degenerate eigenvalue spectrum.
\end{theorem}
The proof of this theorem is given in \ref{Suppl_Thrm_II}. Now, $L^{\rho}(\mathcal{G})$ is a symmetric matrix and can be diagonlized. Consider the diagonalization:
\begin{equation*}
L^{\rho}(\mathcal{G}) = \sum_{i=1}^{N} \gamma^{\rho}_{i}  \sum_{j=1}^{d_{\rho}} \phi^{\rho}_{ij}(\phi^{\rho}_{ij} )^{T}
\end{equation*}
where each $\Phi^{\rho}_{ij} \in \mathbb{C}^{d_{\rho}N}$.
Take the $d_{\rho}$ eigenvectors corresponding to the $d_{\rho}$ algebraically smallest  eigenvalues of $L^{\rho}(\mathcal{G})$ and form the $d_{\rho} \times d_{\rho}N $ matrix
\begin{equation*}
\Phi^{\rho} = \sqrt{N} \begin{bmatrix}
    ( \phi^{\rho}_{11} )^{T} \\
    ( \phi^{\rho}_{12} )^{T} \\
    ... \\
    ( \phi^{\rho}_{1d_{\rho}} )^{T} \\
\end{bmatrix} \in \mathbb{R}^{ d_{\rho} \times d_{\rho} N}
\end{equation*} 
The factor of $\sqrt{N}$ is included for correct normalization, $\Phi^{\rho}(\Phi^{\rho})^{\dagger} = N \cdot \mathbb{I}_{ d_{\rho} }$.

\begin{figure}[h]
	\centering
	\begin{subfigure}{1.0\linewidth}
		\fbox{\includegraphics[width=1.0\textwidth]{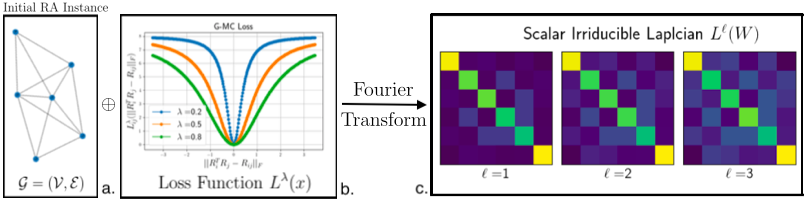}}
	\end{subfigure}
	\caption{ Illustration of $SO(3)$ Multi-Irreducible Robust Synchronization Method. Panel `a' shows a specific graph $\mathcal{G} = ( \mathcal{V} , \mathcal{E} )$ instance of a rotation averaging problem with six nodes. Panel `b' shows the loss function, which is a user input based on risk-preferences and noise model. In panel `c', the Fourier transform on $SO(3)$ is used to compute the scalar irreducible matrices $L^{\ell}(W)$ for $\ell=1,2,3,...$. The relative graph weights of each $L^{\ell}(W)$ are determined by the input graph and the Fourier transform of the loss function. For each $\ell$, the scalar irreducible matrices $L^{\ell}(W)$ are positive definite graph Laplacian matrices, which are frequently studied in spectral graph theory \cite{Chung_1997}.  }
	\vspace{-0.3cm}
\end{figure}

\subsection{ Consensus Reconstruction }\label{Section:Consensus Reconstruction }
\begin{figure}[h]
	\centering
	\begin{tabular}{|l|}
		\hline
		\includegraphics[width=0.49\textwidth]{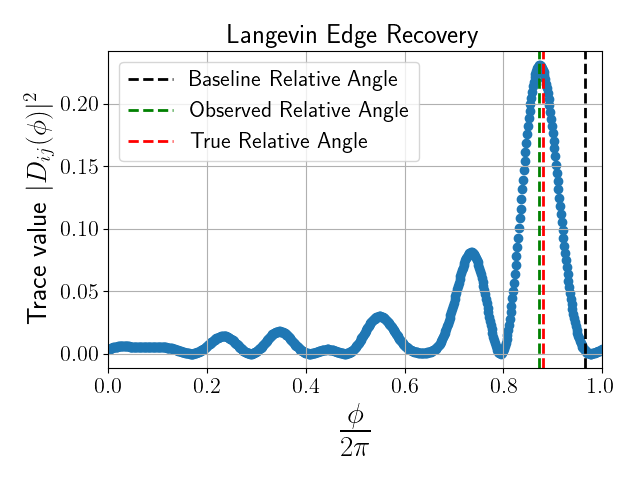}
		\includegraphics[width=0.49\textwidth]{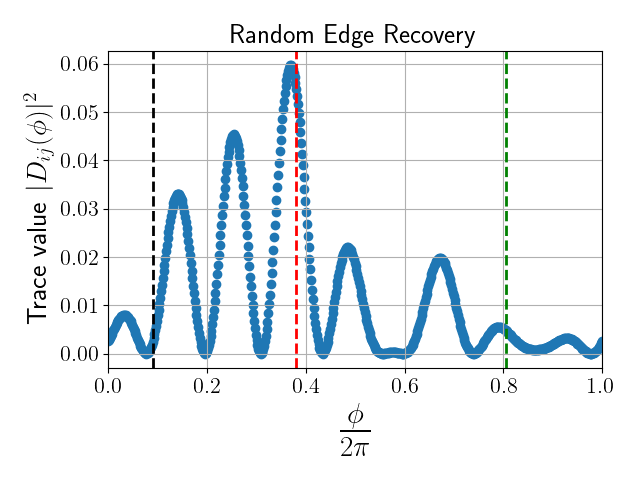} \\
		\hline
	\end{tabular}
	\vspace*{-0.0cm}
	\caption{ $SO(2)$ Edge Recovery. Left panel shows $|D_{ij}(\phi)|^{2}$ as a function of angle $\phi \in S^{1}$ for an edge that is corrupted by Langevin noise. Left panel shows $|D_{ij}(\phi)|^{2}$ as a function of angle $\phi \in S^{1}$ for an edge that is corrupted by totally random noise. The green vertical line denotes the observed difference $\tilde{\phi}_{ij}$. The red vertical line denotes the true angular difference $\phi_{ij}$. The black dotted line denotes the recovered angular difference using the \cite{Doherty_2022} method. Note that while our method and the \cite{Doherty_2022} method recover similar solutions on the edge corrupted by Langevin noise (left), our method is \emph{also} able to recover accurate estimates for edges corrupted by uniform noise (right).    }\label{Table:SO2_Fourier_Distributions}
\end{figure}

Each $\rho$-irreducible sub-problem returns an independent $\Phi^{\rho}$ which is the element of $\text{St}( d_{\rho} , d_{\rho} N )$. We now need to compute the optimal consensus among all sub-problems. This is complicated by the fact that each $\Phi^{\rho}$ may be in a different gauge. However, the term $\Phi^{\rho}_{i}(\Phi^{\rho})^{\dagger}_{j}$ is a gauge invariant quantity that is an estimate of the true $g^{-1}_{i}g_{j}$. We thus can find the consensus on the $g^{-1}_{i}g_{j}$ without dealing with choice of gauge. Specifically, we can compute the inverse Fourier transformation \ref{Fourier Transform on Groups},
\begin{align*}
D_{ij}(g) =  \sum_{ \rho \in G } d_{\rho} \text{Tr}[ \rho(g^{-1}) \Phi_{i}^{\rho}( \Phi_{j}^{\rho} )^{T} ] 
\end{align*}
which gives a function $D_{ij} :SO(d) \rightarrow \mathbb{C}$. We can compute a de-noised set of $\tilde{g}_{ij}$, by computing the maximum of $D_{ij}$,
\begin{align*}
\hat{g}_{ij} = \argmax_{ g \in G }[ \enspace | D_{ij}(g) |^{2} \enspace   ]
\end{align*}
which gives us a de-noised set of $\hat{g}_{ij}$. Note that the computation of $\hat{g}_{ij}$ involves $|\mathcal{E}| \leq \mathcal{O}(N^{2})$ independent single variable optimizations over $G$, as opposed to an optimization over $G^{N} = \mathcal{O}( \exp(N) )$ in the original MLE. In practice we will truncate the Fourier transform at some maximum harmonic $\ell_{\max}$. This optimization over $G$ can be done with sampling methods. Heuristically, as long as the grid discritization over $G$ is less than the Nyquist frequency $ \mathcal{O}(\frac{1}{\ell_{\max}})$ the optimal $g \in G$ will be found (This is formalized in the appendix). A visualization of $|D_{ij}(g)|^{2}$ for $SO(2)$ phase synchronization is shown in \ref{Table:SO2_Fourier_Distributions}. Vertical lines show the true phase, the phase recovered by the spectral method \cite{Doherty_2022}, and the observed phase.

\section{Noise Models}

\subsection{Isotropic Langevin Distribution}

The isotropic Langevin distribution $\text{Lan}( \mathbb{I}_{d} , k )$ in $d$-dimensions with precision parameter $k$ is a probability density with support on $SO(d)$. The probability density for $R \in SO(d)$ is given by,
\begin{align}\label{Langevin Probility Density}
	\text{Pr}[R] dR =	\frac{1}{c_{d}(k)} \exp( k \text{Tr}[R]  ) dR
\end{align}
where $c_{d}(k)$ is the normalization factor,
\begin{align}
	\nonumber	c_{d}(k) = \int_{R\in SO(d)}dR \exp( k \text{Tr}[ R ] ) 
\end{align} 
where the $dR$ denotes integration over the Haar measure \cite{Haar_1933}. This noise model is the maximum entropy distribution over $SO(d)$ given requirements on the distribution mode \cite{Fisher_1953}. The Langevin Distribution is further described in \ref{Suppl_Lang}.

The negative log-likelihood for the Langevin distribution is given by,
\begin{equation*}
-\log p(\tilde{g}_{ij} \mid g_{ij}) = \frac{1}{2} k_{ij}|| D^{1}(g^{-1}_{i}g_{j}) - D^{1}( \tilde{g}_{ij} ) ||^{2}_{F}
\end{equation*}
The computation of the Fourier Transform for this edge loss is given in \ref{Suppl_F_Norm_Loss}. The only non-zero Fourier coefficient is the $\ell=1$ irreducible,
\begin{align*}
\hat{f}^{\ell=1}_{ij} = 
\begin{cases}
&\sum_{j \ne i } k_{ij} \text{ if }i=j \\
& -k_{ij} \text{ if }i \ne j
\end{cases}
\end{align*}
and all other $\hat{f}^{\ell \ne 1} = 0$. Using this choice of Fourier coefficients, our algorithm reduces exactly to \cite{Doherty_2022}.

While many other noise models are possible, the Langevin noise is the maximal entropy distribution on $SO(d)$ and, by the principle of maximum entropy \cite{Jaynes_1957}, is a very natural assumption of noise distribution. We expect that the results we derive in this paper are applicable to a large class of noise models so long as errors on relative measurements $R^{\epsilon}_{ij} = ( R^{\epsilon}_{ji} )^{T}$ are assumed to be uncorrelated and have finite moments.

\subsection{Uniform Random Corruption Model }
A major issue in robotics is that occasionally sensor malfunction will result in measurements that are totally random \cite{Rosen_2021,Cadena_2016}. A failed sensor on the $ij$-th edge can be modeled by the noise term
\begin{align*}
R^{\epsilon}_{ij} \sim \text{Uniform}[ SO(d) ]
\end{align*}
so that the error term $R^{\epsilon}_{ij}$ is sampled uniformly on $SO(d)$. We show that our proposed method can deal with random noise corruption through the use of more general M-estimators, which are loss functions that do not arise as negative log-likelihoods of a generative model. Specifically, we are free to chose the $ij$-th edge-loss as 
\begin{align*}
f_{ij}(g_{i},g_{j}; \tilde{g}_{ij} ) = \mathbf{L}^{\lambda}_{ij}( || g^{-1}_{i}g_{j} - \tilde{g}_{ij} ||_{F}  )
\end{align*}
where $\mathbf{L}_{ij}^{\lambda}: \mathbb{R}^{+} \rightarrow \mathbb{R}^{+}$ is a loss function parameterized by the (user imput) $\lambda$. We will specifically consider the Cauchy-loss and German-McClure (G-MC) loss, but our proposed method allows for \emph{any} edge-loss function.
\begin{align*}
& \text{Cauchy Loss: }\mathbf{L}^{\lambda}(x) = \frac{\lambda^{2}}{2}\log( 1 + (\frac{x}{\lambda})^{2} )\\
& \text{German-McClure (G-MC) Loss: }\mathbf{L}^{\lambda}(x) = \frac{2x^{2}}{x^{2}+4\lambda^{2}}
\end{align*}
The computation of the Fourier Transforms of these functions in $SO(2)$ and $SO(3)$ are shown in \ref{Table:SO3_Fourier_Transform_Losses}. The Fourier Transform on groups for the robust loss functions $L^{\lambda}$ generate a sequence of $\lambda$ dependent coefficients $K^{\rho}(\lambda)$, one for each $G$-irreducible $\rho$. For the $ij$-th edge, the loss function $\lambda$ parameter $\lambda_{ij}= \lambda_{ij}(k_{ij})$ is chosen as a function of the Langevin concentration parameter $k_{ij}$. Thus, for each edge, the Fourier Transform on groups generate a sequence of coefficients $K^{\rho}(\lambda_{ij}(k_{ij}) )$. In practice, we choose $\lambda_{ij}$ so that the $L^{\lambda}$ deviation from quadratic occurs at roughly three standard deviations of a Langevin random variable with concentration parameter $k_{ij}$. This is discussed in depth in \ref{Suppl_Lang_Fit}.

\section{Noiseless Case}

The relaxed MLE problem \eqref{Relaxed MLE} is a matrix eigenvalue problem. In order to derive bounds on the error of \eqref{Relaxed MLE} with random noise, we will first study the noiseless case. Consider the problem instance where all $\tilde{g}_{ij} = g_{ij}$. We will define the noiseless rotation Laplacian of the true $g_{ij}$ as 
\begin{align}\label{Definition of Noiseless Laplcian Matrix}
L^{\rho}(G)_{ij} \equiv 
\begin{cases}
& D^{\rho}_{i} \mathbb{I}_{d_{\rho}} \text{ if } i=j \\
& - K^{\rho}_{ij}\rho(g_{ij}) \text{ if } i \ne j
\end{cases}
\end{align}	
which corresponds to the matrix \eqref{Definition of Noisy Laplcian Matrix} with each measured relative rotation equal to the true relative rotation. In the noiseless case, $L^{\rho}(G)$ can be transformed into a much simpler matrix. Define $S_{\rho} \in \mathbb{R}^{d_{\rho}N \times d_{\rho}N }$ to be the $d_{\rho} \times d_{\rho}$ block diagonal matrix
\begin{align}\label{S Matrix}
S_{\rho} = \begin{bmatrix}
	\rho(g_{1}) & \mathbb{O}_{d_{\rho}} & .... & \mathbb{O}_{d_{\rho}} \\
	\mathbb{O}_{d_{\rho}} & \rho(g_{2}) & ... & \mathbb{O}_{d_{\rho}}\\
	... & ... & ... & ...\\\
	\mathbb{O}_{d_{\rho}} &\mathbb{O}_{d_{\rho}} & ... & \rho(g_{N}) \\
\end{bmatrix}
\end{align}
If we conjugate \eqref{Definition of Noiseless Laplcian Matrix} by $S_{\rho}$, we have that
\begin{equation*} 
	\nonumber	S_{\rho} L^{\rho}(G) S_{\rho}^{-1} = L^{\rho}(W) \otimes \mathbb{I}_{ d_{\rho} }
\end{equation*}
where $ L(W) $ is an $N \times N$ matrix with matrix elements given by
\begin{align}\label{Weight Laplacian}
L^{\rho}(W)_{ij} = \begin{cases}
 &   D^{\rho}_{i} \text{ for } i=j\\
 & - K^{\rho}_{ij} \text{ for }i \ne j
\end{cases}
\end{align}
Thus, the spectrum of the noiseless rotation $\rho$-Laplacian \eqref{Definition of Noiseless Laplcian Matrix} is the spectrum of $L^{\rho}(W)$ with $d_{\rho}$-fold multiplicity. When all $K^{\rho}_{ij} \geq 0$ (which is always true for both the Cauchy and G-MC losses, see \ref{Table:SO3_Fourier_Transform_Losses} ), properties of matrices of the form $L^{\rho}(W)$ have been thoroughly studied in graph theory \cite{Chung_1997}. For a connected graph, $L^{\rho}(W)$ is positive semi-definite with lowest eigenvalue equal to zero. Furthermore, the second smallest eigenvalue of $L^{\rho}(W)$ is strictly positive if the graph is connected. Furthermore, the lowest eigenvalue of $L^{\rho}(W)$ is always of the form
\begin{align*}
\phi = \frac{1}{\sqrt{N}}\begin{bmatrix}
1 \\
1\\
...\\
1
\end{bmatrix}
\end{align*}
Thus, the eigenspace corresponding to the $d_{\rho}$-degenerate smallest eigenvalues of $L^{\rho}( \mathcal{G} ) $ is spanned by the vectors
\begin{align*}
S_{\rho}( \phi \otimes e_{1} ), \enspace S_{\rho}( \phi \otimes e_{2} ), \enspace ... \enspace, S_{\rho}( \phi \otimes e_{d_{\rho}} )
\end{align*}
Thus, the $ij$-th block of the outer product $\Phi^{\rho} (\Phi^{\rho})^{\dagger}$ is given by
\begin{align*}
[ \Phi^{\rho} (\Phi^{\rho})^{\dagger} ]_{ij} = N \phi_{i} \phi_{j} \rho( g_{ij} )  = \rho(g_{ij})
\end{align*}
Taking the inverse Fourier transform, we have that
\begin{align*}
D_{ij}(g) = \sum_{ \rho \in \hat{G} } d_{\rho} \text{Tr}[ \rho(g^{-1})  \rho( g_{ij} )   ] = \delta_{g_{ij}}(g)
\end{align*}
So $D_{ij}(g)$ is exactly the Fourier transform of the delta function centered at $g_{ij}$ \ref{Delta Function Fourier Transform on Groups}. Ergo, in the noiseless case,
\begin{align*}
g_{ij} = \argmax_{g \in G} |D_{ij}(g)|^{2} 
\end{align*}
holds exactly and the recovery produces the exact solution. In the noisy case, the recovered $\tilde{D}_{ij}(g)$ will not be exactly a delta function centered at $g_{ij}$. However, the Parsival-Plancheral theorem (see \ref{Suppl_Peter-Weyl} for details) bounds the deviation from the noiseless case. Specifically,
\begin{align*}
|| \tilde{D}_{ij} - D_{ij} ||^{2}_{ L^{2}(G) } = \sum_{\rho \in \hat{G} } d^{2}_{\rho} || \tilde{D}_{ij}^{\rho} - D^{\rho}_{ij} ||^{2}_{F}
\end{align*}
In the case of Langevin noise, it is possible to derive probabilistic bounds on the term $|| \tilde{D}_{ij}^{\rho} - D^{\rho}_{ij} ||^{2}_{F}$ using an argument similar to that of \cite{Doherty_2022}. Thus, bounds on the deviation of the recovered eigenvectors of $L^{\rho}(\mathcal{G})$ from their true values translate into bounds on the recovered $g_{ij}$. It should be noted that
\begin{align*}
 \forall g\in G, \enspace |D_{ij}(g)|^{2} \geq 0, \quad \int_{g \in G}dg \text{ }  |D_{ij}(g)|^{2} = 1
\end{align*}
so that $|D_{ij}(g)|^{2}$ can be interpreted as a probability distribution on $G$.

\begin{figure*}[h!]
\begin{subtable}{1.00\linewidth}
\centering
\begin{tabular}{|c|c|c|c|c|c|c|}
\hline
& \multicolumn{6}{c|}{ Percent of Edges Corrupted }    \\  
\hline
Method & 0.00 & 0.01  & 0.05   & 0.1   & 0.15 & 0.20    \\
\hline
\textbf{Baseline Methods} & &  &  &  &  &    \\
Spectral Relaxation \cite{Doherty_2022}   & 0.04(0.09) & 0.27(0.31) & 0.40(0.42) & 0.49(0.61) & 0.41(0.74) & 0.55(0.81)   \\
Weiszfeld Algorithm \cite{Hartley_2011_L1}   & 0.03(\textbf{0.05}) & \textbf{0.15} (0.28) & 0.25(0.38) & 0.34 (0.43) & 0.38(0.54) &  0.46(0.61) \\
Shonan Rotation Averaging \footnote{Implemented using GTSAM: https://gtsam.org/doxygen/a01058.html } \cite{Dellaert_2020_Shonan} & \textbf{0.02} (0.07) & 0.25 (0.31) & 0.36 (0.41) & 0.45 (0.52) & 0.443(0.57) & 0.55 (0.76)  \\
Graduated Non-Convexity \footnote{Implemented using GTSAM:https://github.com/borglab/gtsam/blob/develop/gtsam/nonlinear/GncOptimizer.h } \cite{Yang_2020_Graduated} & 0.04(0.08) & \textbf{0.15}(0.31) & 0.21(0.35) & 0.31(0.44) & 0.34(0.58) & 0.39(0.83) \\
\hline
\hline
\textbf{Cauchy Loss (Ours)} & &  &  &  &  &  \\
Irreducible Order $\ell=3$ & 0.12(0.16)& 0.31(0.38) & 0.41(0.46)  & 0.45(0.49) & 0.50(0.56) & 0.45(0.72)  \\
Irreducible Order $\ell=5$ & 0.06(0.11) & 0.28(0.31) & 0.35(0.37)  & 0.36(0.43) & 0.38(\textbf{0.54}) & 0.42(0.74 )  \\
Irreducible Order $\ell=8$& 0.05(0.08) & 0.16(\textbf{0.25}) & \textbf{0.15}(\textbf{0.31})  & 0.28(0.45) & \textbf{0.30}(\textbf{0.54} ) & 0.35(\textbf{0.59})  \\
\hline
\hline
\textbf{GMC Loss (Ours)} & &  &  &  &  &   \\
Irreducible Order $\ell=3$ & 0.11(0.17) & 0.27(0.39) & 0.36(0.48)  & 0.38(0.59) & 0.39(0.71 )  & 0.41(0.83)  \\
Irreducible Order $\ell=5$ & 0.09(0.11)& 0.19(0.32) & 0.31(0.42)  & 0.33(0.47 ) & 0.35(\textbf{0.54}) & 0.39(0.63) \\
Irreducible Order $\ell=8$ & 0.06(0.08) & 0.15(0.29) & 0.23(0.36)  & \textbf{0.27}(\textbf{0.39}) & 0.29(0.55) & \textbf{0.32} (0.60)  \\
\hline
\end{tabular}
\end{subtable}
\caption{Comparison of $SO(3)$ robust rotation averaging methods for different methods on the sphere dataset \cite{Carlone_2015_Lagrangian}. The normalized $F$-norm loss $d_{F}$ is shown outside of parenthesis, the normalized $d_{\infty}$ norm is show in parenthesis. Shonan rotation averaging \cite{Dellaert_2020_Shonan} and the Spectral Relaxation method \cite{Doherty_2022} are designed for data with Langevin noise. The Wiezfeld Algorithm \cite{Hartley_2011_L1} and Graduated Non-Convexity \cite{Huang_2016} are specifically designed as robust rotation averaging methods. All method except \cite{Doherty_2022} and ours involve multiple iterative updates. Each datapoint is averaged over $10$ trials. Additional numerical experiments are shown in appendix \ref{Numerical Experiments on so3 Rotation Averaging}  }\label{Main:Table:3d_RA_corrupt}
\end{figure*}

\section{Distance Metrics}\label{Section:Distance Metrics}

The spectral algorithm will produce an MLE estimate $\hat{R}_{i}$. We would like to measure how similar the recovered solution $\hat{R}_{i}$ is to the ground truth $R_{i}$. In order to do so, we will need to define distance metrics on vectorizations of $N$ $SO(d)$ elements. This problem is slightly complicated by the fact that a solution to the rotation averaging problem is only defined up to gauge transformation by left multiplication of an element of $R \in SO(d)$ \eqref{Gauge Invarience} \cite{Rosen_2019}. Therefore, it is necessary to define notions of distance that account for the presence of this $SO(3)$-gauge symmetry.  

Let $R = [R_{1},R_{2},...,R_{N}]$ and $\hat{R} = [\hat{R}_{1},\hat{R}_{2},...,\hat{R}_{N}]$ be vectorizations of $N$ group elements. We can then define a gauge invariant distance metrics over $SO(3)$ as follows:
\begin{align}\label{Standard Distances}
& d_{F}( R , \hat{R} ) \triangleq \min_{ U \in SO(d) }|| \rho( U \circ R )  - \rho( R ) ||_{F}
\end{align}
where $U \circ g = [ UR_{1},UR_{2}, ... , UR_{N} ]$ denotes the gauge transformation of left multiplication by $U$. There is a closed form expression for the optimal minimizer $U$ in terms of $R$ and $\hat{R}$ \cite{Doherty_2022}. The distance metrics defined in \ref{Standard Distances} are \emph{global} error measurements. We can also consider a \emph{local} error metric, which measures
\begin{align}\label{Local Distances}
&	d_{\infty}( R , \hat{R} ) \triangleq \min_{U \in SO(d) }[ \max_{1\leq i \leq N}[ || U R_{i}  -  \hat{R}_{i} ||_{F} ] ] 
\end{align}
The \ref{Local Distances} metrics are more difficult to work with, but are more important in real robotics applications. There is no closed form for the gauge choice in \ref{Local Distances}, but the optimal $U$ in \ref{Standard Distances} serves as a good proxy. The $F$-norm error \ref{Standard Distances} may have a small loss if all errors are concentrated in a single estimation of a rotation, the error metric in \ref{Local Distances} is a worst case quantity that can only be small if each rotation estimation is accurate.
The distance metrics we work with satisfy the bounds \cite{Rosen2016Certifiably}
\begin{align*}
& 0 \leq  d_{F}( R, \hat{R}) \leq \ 2\sqrt{d N}, \\
& 0 \leq  d_{\infty}(R, \hat{R}) \leq 2\sqrt{d}
\end{align*}
We will work with $d_{F}$ and $d_{\infty}$ normalized to be between zero and one.

\section{Baselines}

We benchmark our algorithm on a set of commonly used rotation averaging datasets \cite{Carlone_2015_Lagrangian}. We compare our algorithm with some standard rotation averaging baselines. Specifically, we compare our method with the spectral relaxation proposed in \cite{Doherty_2022} and the Shonan rotation averaging method \cite{Dellaert_2020_Shonan}, which is a certifiable rotation averaging method. We also compare our method with the Weiszfeld algorithm \cite{Hartley_2011_L1} and the Graduated Non-Convexity (GNC) method \cite{Yang_2020_Graduated}, which are specifically designed to be robust to outlier contamination. Our method is able to produce accurate reconstructions, even with high outlier contamination.

\section{Conclusion}

In this work, we have show that method inspired by cryoEM \cite{Gao_2019_Multifrequency, Bandeira_2015_Nonunique} can be applied to the $SO(3)$ rotation averaging problem in computer vision. Unlike \cite{Gao_2019_Multifrequency,Bandeira_2015_Nonunique} we use a single simple spectral relaxation \cite{Doherty_2022} in our method. Using the Fourier Transform on groups, our method can be used for rotation averaging with any loss function. We empirically show that adding multiple non-equivalent copies of $SO(3)$ representations for rotation averaging is significantly more robust to noise. Our method can be seen as a representation theoretic extension of \cite{Doherty_2022}, which reduces to \cite{Doherty_2022} with quadratic loss. We consider a few different quantification of closeness, corresponding to the $F$-norm and $\infty$-norm. Our results will be especially useful for SLAM-type problems and will have applications to fields such as robotic grasping, remote sensing and autonomous driving.

\subsection{Limitations}
The main limitation of our method is the time cost of the consensus reconstruction step \ref{Section:Consensus Reconstruction }. However, each edge optimization in the consensus reconstruction is independent, and each optimization can be computed in parallel (please see our code release for open source implementation).

\subsection{Future Work}
In many computer vision applications the relative rotations between poses must be estimated from data. Our method suggests that group theory plays an intimate role in designing noise resistant pose estimation algorithms, and implies that equivarient machine learning \cite{Bronstein_2021_Geometric} is a natural method for solving rotation averaging problems. Specifically, $SO(3)$/$SE(3)$-equivariant graph neural networks seem ideally suited for tackling the RA/SLAM problem. Furthermore, the proposed multi-irreducible synchronization method can be applied to synchronization over \emph{any} compact group. Point correspondence problems in computer vision can be written as a synchronization problem over the symmetric group on $n$ elements \cite{Pachauri_2013_Solving}.


\small
\section{Acknowledgments and Disclosure of Funding}
Owen Howell thanks Dr. Matt Giamou, David Klee and Haojie Huang for useful discussions. Owen Howell further thanks Shankara Narayanan for code implementation advice. Owen Howell acknowledges support from the National Science Foundation Graduate Research Fellowship Program (NSF-GRFP) and the Yochiro Nambu Fellowship from the University of Chicago.
\normalsize

\bibliographystyle{plain}
\bibliography{refs}

\appendix

\newpage
\clearpage
\setcounter{page}{1}
\appendix
\onecolumn

 \section{Proof of Theorem I}\label{Suppl_Thrm_I}
Let $h : \mathbb{R}^{+} \rightarrow \mathbb{R}^{+}$ be a smooth function. Let $(\sigma , V)$ be any representation of the group $G$. Let $|| \cdot ||$ be a unitary norm on the vector space $V$. Consider a loss function of the form
\begin{align*}
	f(g) = h( || \sigma(g) - \sigma(g') || )
\end{align*}
where $|| \cdot ||$ is a unitary norm and $h: \mathbb{R}^{+} \rightarrow \mathbb{R}^{+}$. We can compute the Fourier transform of $f$ with respect to the $\rho \in \hat{G}$ irreducible,
\begin{align*}
\hat{f}^{\rho} = \int_{g\in G}dg\text{ } f(g) \rho(g) = \int_{g\in G}dg\text{ }  h( || \sigma(g) - \sigma(g') || ) \rho(g) 
\end{align*}
making the change of variables $g \rightarrow g'g $, the integral reduces to 
\begin{align*}
\hat{f}^{\rho} = \int_{g\in G}dg\text{ }  h( || \sigma(g'g) - \sigma(g') || )\rho(g'g)
\end{align*}
Now, using the definition of group representations, $\rho(g'g) = \rho(g')\rho(g)$ and $\sigma(g'g) = \sigma(g')\sigma(g)$. Furthermore, the norm $||\cdot||$ is unitary and  $|| \sigma(g'g) - \sigma(g')|| = ||\sigma(g')( \sigma(g) - \mathbb{I}_{d_{\sigma}} )|| = ||( \sigma(g) - \mathbb{I}_{d_{\sigma}} )||$. Ergo,
\begin{align*}
\hat{f}^{\rho} = \rho(g') \int_{g\in G}dg\text{ }  h( || \sigma(g) - \mathbb{I}_{d_{\sigma}} || )\rho(g)
\end{align*}
Note that the remaining integral is independent of $g'$. Now consider the integral
\begin{align*}
	C^{\rho} = \int_{g\in G}dg\text{ }  h( || \sigma(g) - \mathbb{I}_{d_{\sigma}} || )\rho(g) 
\end{align*}
Let $h$ be any element of $G$. We can left multiply $C^{\rho}$ by $\rho(h)$,
\begin{align*}
\forall h \in G, \quad \rho(h) C^{\rho} = \int_{g\in G}dg\text{ }  h( || \sigma(g) - \mathbb{I}_{d_{\sigma}} || )\rho(hg) 
\end{align*}
Now, making the normal transformation, $g \rightarrow h^{-1}gh$, we have that
\begin{align*}
\forall h \in G, \quad \rho(h) C^{\rho} = \int_{g\in G}dg\text{ }  h( || \sigma(h^{-1}gh) - \mathbb{I}_{d_{\sigma}} || )\rho(gh) 
\end{align*}
and again using the unitarity of the $|| \cdot ||$ norm, we have that
\begin{align*}
\forall h \in G, \quad \rho(h) C^{\rho} = \int_{g\in G}dg\text{ }  h( || \sigma(g) - \mathbb{I}_{d_{\sigma}} || )\rho(gh) 
\end{align*}
Now, using the relation $\rho(gh) = \rho(g)\rho(h)$, we have that, for all irriducibles $\rho \in \hat{G}$,
\begin{align*}
\forall g\in G, \quad \rho(g) C^{\rho} = C^{\rho} \rho(g) 
\end{align*}
Thus, by Schur's lemma \ref{Suppl_Schur_Lemma}, we have that $C^{\rho} = K^{\rho} \mathbb{I}_{d_{\rho}}$ where $K^{\rho} \in \mathbb{C}$ is a constant. Using a corollary of Schur's lemma \ref{Suppl_Schur_Lemma}, we have that
\begin{align*}
C^{\rho} = \frac{ \text{Tr}[C^{\rho}]  }{d_{\rho}} \mathbb{I}_{d_{\rho}} = \frac{1}{d_{\rho}} \text{Tr}[ \enspace \int_{g\in G}dg\text{ }  h( || \sigma(g) - \mathbb{I}_{d_{\sigma}} || )\rho(g) \enspace ] \mathbb{I}_{d_{\rho}}
\end{align*}
Thus, the constant of proportionality is given by
\begin{align*}
K^{\rho} = \frac{1}{d_{\rho}} \text{Tr}[ \enspace \int_{g\in G}dg\text{ }  h( || \sigma(g) - \mathbb{I}_{d_{\sigma}} || )\rho(g) \enspace ] 
\end{align*}

To summarize, let $h : \mathbb{R}^{+} \rightarrow \mathbb{R}^{+}$ be a smooth function. Let $(\sigma , V)$ be any representation of the group $G$. Let $|| \cdot ||$ be a unitary norm on the vector space $V$. Consider a loss function of the form
\begin{align*}
	f(g) = h( || \sigma(g) - \sigma(g') || )
\end{align*}
Then, the Fourier transform with respect to the $\rho$-irreducible is given by
\begin{align*}
\hat{f}^{\rho} = K^{\rho} \rho(g') 
\end{align*}
where $K^{\rho} \in \mathbb{C}$ is a scalar constant independent of $g'$ given by
\begin{align*}
K^{\rho} = \frac{ \text{Tr}[K^{\rho}]  }{d_{\rho}} \mathbb{I}_{d_{\rho}} = \frac{1}{d_{\rho}} \text{Tr}[ \enspace \int_{g\in G}dg\text{ }  h( || \sigma(g) - \mathbb{I}_{d_{\sigma}} || )\rho(g) \enspace ] 
\end{align*}
Interchanging integration and trace, we have that
\begin{align*}
K^{\rho} = \frac{1}{d_{\rho}}  \int_{g\in G}dg\text{ }  h( || \sigma(g) - \mathbb{I}_{d_{\sigma}} || )\text{Tr}[\rho(g) ] = \frac{1}{d_{\rho}}  \int_{g\in G}dg\text{ }  h( || \sigma(g) - \mathbb{I}_{d_{\sigma}} || ) \chi^{\rho}(g)
\end{align*}
where $\chi^{\rho}(g) = \text{Tr}[ \rho(g) ]$ is the character of the irreducible $\rho$ representation. This integral over the Haar-mesure is invariant under the normal transformation $g \rightarrow hgh^{-1}$. Integrals of this form can be further simplified using the Weyl integration formula \cite{Weyl_1926}. Heuristically, the Weyl formula allows one to evaluate integrals on compact non-commutative groups in terms of integrals over the largest commutative subgroup of $G$. 

\subsection{Weyl Integration Formula in $SO(3)$}
We briefly review the Weyl Integration formula in $SO(3)$\cite{Boumal_2022}. Let $f : SO(3) \rightarrow \mathbb{C} $ be a complex valued function on $SO(3)$ that is \emph{normal} with $f(hgh^{-1}) = f(g)$ for all $h,g\in SO(3)$. The Weyl formula in $SO(3)$ states that
\begin{align*}
\int_{g \in SO(3)}dg \text{ }f(g) = \int_{0}^{2\pi}d\phi \text{ } [1 - \cos(\phi) ] f( g_{\phi} )  
\end{align*}
where $g_{\phi}$ is a rotation of angle $\phi$ about the $z$-axis. Thus, using the Weyl Integration Formula in $SO(3)$ in \ref{Suppl_Thrm_I}, we have that
\begin{align*}
K^{\ell} = \frac{1}{2\ell+1}  \int_{0}^{2\pi}d\phi\text{ } [1-\cos(\phi)] h( || \sigma(g_{\phi}) - \mathbb{I}_{d_{\sigma}} || )\chi^{\ell}(g_{\phi})
\end{align*}
Now, the characters of $SO(3)$ representations are given by
\begin{align*}
\chi^{\ell}( g_{\phi} ) = \frac{ \sin( (\ell + \frac{1}{2}) \phi  )  }{ \sin( \frac{1}{2}\phi ) } = \sum_{k=-\ell}^{\ell} \exp(ik\phi)
\end{align*}
and the character of the $\ell$-th $SO(3)$ irreducible evaluated on a rotation of angle $\phi$ about the $z$-axis is the Dirichlet kernel of order $\ell$ evaluated at $\phi$. We thus have that
\begin{align*}
K^{\ell} = \frac{1}{2\ell+1}  \int_{0}^{2\pi}d\phi\text{ } [1-\cos(\phi)] h( || \sigma(g_{\phi}) - \mathbb{I}_{d_{\sigma}} || ) \frac{ \sin( (\ell + \frac{1}{2}) \phi  )  }{ \sin( \frac{1}{2}\phi ) }
\end{align*}
For loss functions that depend only on the $\ell=1$ fundamental representation, we have that $|| D^{1}(g) - \mathbb{I}_{3} ||^{2}_{F} = 6 - 2\chi^{1}(g)  $
\begin{align*}
K^{\ell} = \frac{1}{2\ell+1}  \int_{0}^{2\pi}d\phi\text{ } [1-\cos(\phi)]   \frac{ \sin( (\ell + \frac{1}{2}) \phi  )  }{ \sin( \frac{1}{2}\phi ) } h( \sqrt{ 4 - 4 \cos(\phi) } )
\end{align*}
This is an integral over a single variable in the range $[0,2\pi]$ and can easily be evaluated with numerical methods.

\subsection{Cesàro Summation and Fejér Kernels}\label{Cesàro Summation and Fejér Kernels}

The expression for $K^{\ell}$ involves an expression over the Dirichlet kernel, and is not guaranteed to be positive definite. However, using the results of \cite{Bandeira_2015_Nonunique}, it is possible to use second order Cesàro summation for faster convergence. The second order Cesàro sum of the Dirchlet Kernel $D^{\ell}(\phi)$ is the Fejér kernel $F^{\ell}(\phi)$, defined as
\begin{align*}
F^{\ell}(\phi) = \frac{1}{\ell} \sum_{k=0}^{\ell - 1} D^{k}(\phi) = \frac{1}{\ell} \frac{1 - \cos(\ell \phi )}{1 - \cos(\phi) }
\end{align*}
The Fejér kernel is positive definite $F^{\ell}(\phi)\geq 0$. Thus, when using Fejér kernels, the resultant expression for the $K^{\ell}$ are given by
\begin{align*}
K^{\ell} = \frac{1}{2\ell+1}  \int_{0}^{2\pi}d\phi\text{ } [1-\cos(\phi)] F^{\ell}(\phi)  h( \sqrt{ 4 - 4 \cos(\phi) } ) = \frac{1}{\ell(2\ell + 1)}  \int_{0}^{2\pi}d\phi\text{ } [1-\cos(\ell \phi)]  h( \sqrt{ 4 - 4 \cos(\phi) } )
\geq 0 \end{align*}
so that the resulting coefficients $K^{\ell} \geq 0$ are positive definite for all loss functions $h( \cdot ) \geq 0$. Furthermore, the expression for $K^{\ell}$ is an integral over a single variable in the range $[0,2\pi]$ and can easily be evaluated with numerical methods.

\section{Proof of Theorem II}\label{Suppl_Thrm_II}
Let $h : \mathbb{R}^{+} \rightarrow \mathbb{R}^{+}$ be a smooth function. Let $(\sigma , V)$ be any representation of the group $G$. Let $|| \cdot ||$ be a unitary norm on the vector space $V$. Suppose that the edge-loss functions take the form
\begin{align*}
	f_{ij}(g_{i},g_{j};\tilde{g}_{ij}) = h_{ij}( || \sigma(g^{-1}_{i}g_{j}) - \sigma(\tilde{g}_{ij}) || )
\end{align*}
where $|| \cdot ||$ is a unitary norm and $h_{ij}: \mathbb{R}^{+} \rightarrow \mathbb{R}^{+}$. Using \ref{Suppl_Thrm_I}, the Fourier transform of each edge loss function is given by
\begin{align*}
    \hat{f}_{ij}^{\rho} = K^{\rho}_{ij} \rho( \tilde{g}_{ij} )
\end{align*}
where $K^{\rho}_{ij} \in \mathbb{C}$. Let $L^{\rho}( \mathcal{G} )$ be the $\rho$ irreducible Laplacian weight matrix with $d_{\rho} \times d_{ \rho } $ blocks given by
\begin{align*}
L^{\rho}(\mathcal{G})_{ij} = K^{\rho}_{ij} \rho( \hat{g}_{ij} )
\end{align*}
Now, the true solution is only defined up to left multiplication by an element of $g \in G$. Thus, left multiplication by the block diagonal matrix
\begin{align*}
\sigma_{\rho}(g) = \begin{bmatrix}
    \rho(g) & 0 & ...& 0 \\
     0& \rho(g) & ...& 0 \\
     ... & ... & ...& ... \\
     0 & 0 & 0 & \rho(g) \\
\end{bmatrix}
\end{align*}
must commute with the matrix $L^{\rho}( \mathcal{G} )$ for all $g\in G$,
\begin{align*}
 \forall g\in G, \quad \sigma_{\rho}(g) L^{\rho}(\mathcal{G}) = L^{\rho}(\mathcal{G}) \sigma_{\rho}(g) 
\end{align*}
Ergo, the matrix $L^{\rho}(\mathcal{G})$ is an element of the vector space
\begin{align*}
L^{\rho}(\mathcal{G}) \in \Hom_{G}[ \bigoplus_{i=1}^{N} (\rho , V) , \bigoplus_{i=1}^{N} (\rho , V) ] \cong \mathbb{C}^{N \times N}
\end{align*}
where the last equality follows from the extended Schur lemma \ref{Suppl_Extended_Schur_Lemma}. Using the \ref{Suppl_Extended_Schur_Lemma}, any $ L^{\rho}(\mathcal{G}) \in \Hom_{G}[ \bigoplus_{i=1}^{N} (\rho , V) , \bigoplus_{i=1}^{N} (\rho , V) ]$ can be written uniquely as
\begin{align*}
L^{\rho}(\mathcal{G}) = U \begin{bmatrix}
\Phi_{11} \mathbb{I}_{ d_{\rho}  } & \Phi_{12} \mathbb{I}_{ d_{\rho}  } & ... & \Phi_{1N} \mathbb{I}_{ d_{\rho}  } \\
\Phi_{21} \mathbb{I}_{ d_{\rho}  } & \Phi_{22} \mathbb{I}_{ d_{\rho}  } & ... & \Phi_{2N} \mathbb{I}_{ d_{\rho}  } \\
... & ... & ... & ... \\
\Phi_{N1} \mathbb{I}_{ d_{\rho}  } & \Phi_{N2} \mathbb{I}_{ d_{\rho}   } & ... & \Phi_{NN} \mathbb{I}_{ d_{\rho}}   \\
\end{bmatrix} U^{\dagger}
\end{align*}
where $U$ is a fixed unitary matrix and each $\Phi_{ij} \in \mathbb{C}$ is a complex scalar. Using tensor product notation, we have that
\begin{align*}
L^{\rho}(\mathcal{G}) = U [ \begin{bmatrix}
\Phi_{11}  & \Phi_{12}  & ... & \Phi_{1N} \\
\Phi_{21}  & \Phi_{22}  & ... & \Phi_{2N}  \\
 ... & ... & ... & ... \\
 \Phi_{N1}  & \Phi_{N2} & ... & \Phi_{NN}    \\
\end{bmatrix} \otimes \mathbb{I}_{ d_{\rho}  } ] U^{\dagger} = U[ \Phi^{\rho}( \mathcal{G} ) \otimes \mathbb{I}_{d_{\rho}} ] U^{\dagger}
\end{align*}
where the matrix $\Phi^{\rho}( \mathcal{G} )$ has elements $\Phi^{\rho}( \mathcal{G} ) = \Phi_{ij}$. Note that the Hermicity of $[ L^{\rho}( \mathcal{G}) ]^{\dagger} = L^{\rho}( \mathcal{G}) $ implies that $ [\Phi^{\rho}( \mathcal{G} ) ]^{\dagger} = \Phi^{\rho}( \mathcal{G} ) $. Thus, the spectrum of the matrix $L^{\rho}(\mathcal{G})$ is the spectrum of the matrix $\Phi^{\rho}( \mathcal{G} )  \otimes \mathbb{I}_{d_{\rho}}$, and has a $d_{\rho}$-degenerate eigenvalue spectrum.

\section{Notation and Preliminaries}\label{Section:Notation and Preliminaries}
To begin, we will fix some notation, trying to stay as close to convention as possible. 

\subsection{Linear Algebra}\label{Section:Linear Algebra}

For any vector $v \in \mathbb{R}^{n}$, the standard 2-norm and $\infty$-norms are defined as 
\begin{align}
	& v \in \mathbb{R}^{n} , \quad || v ||_{2} = ( \sum_{i=1}^{n} v_{i}^{2} )^{\frac{1}{2}} , \quad || v ||_{\infty} = \max_{i\in1,2,...N} |v_{i}|
\end{align}
The $2$-norm and $\infty$-norm are related by
\begin{align}\label{Two Norm and Infinity Norm Relations}
	|| v ||_{\infty} \leq || v ||_{2} \leq \sqrt{n} || v ||_{\infty}
\end{align}

The identity matrix in $d$-dimensions with be denoted as $\mathbb{I}_{d}$. The $d \times d$ matrix of all zeros will be denoted as $\mathbb{O}_{d}$. The set of standard Euclidean basis vectors in $k$-dimensions will be denoted as $\{ e_{i} \}_{i=1}^{k}$. They are given by
\begin{align}\label{Standard Eucldian Basis}
	e_{1} = \begin{bmatrix}
		1\\
		0\\
		0\\
		...\\
		0
	\end{bmatrix}, \quad e_{2} = \begin{bmatrix}
		0\\
		1\\
		0\\
		...\\
		0
	\end{bmatrix}, \quad e_{3} = \begin{bmatrix}
		0\\
		0\\
		1\\
		...\\
		0
	\end{bmatrix}, \quad, ... , e_{k} = \begin{bmatrix}
		0\\
		0\\
		0\\
		...\\
		1
	\end{bmatrix}
\end{align}
The set of $\{e_{i} \}_{i=1}^{k}$ form an orthonormal basis of $\mathbb{R}^{k}$. They satisfy $e^{T}_{i}e_{j}= \delta_{ij}$. The tensor product space is the space spanned by the basis $\{	e_{i} \otimes e_{j} \}_{ij}$. For any matrix $X\in \mathbb{R}^{n \times k} $,  the spectral norm $|| \cdot ||_{2}$ is defined as 
\begin{align}
	& X \in \mathbb{R}^{n \times k} , \quad || X ||_{2} = \max_{ v \in \mathbb{R}^{k} , ||v||_{2} =1 }  || Xv ||_{2} 
\end{align}

The matrix infinity-norm $|| \cdot ||_{\infty}$ is given by
\begin{align}\label{Definition of Infinity Norm}
	& X \in \mathbb{R}^{n \times k} , \quad || X ||_{\infty} = \max_{ i \in 1,2,...,n } \sum_{j=1}^{k} |X_{ij}|
\end{align}
which is the maximum of the sum of the absolute values of the rows.

Similarly, let $X^{T} = X \in \mathbb{R}^{n \times n }$ be a symmetric matrix. We will denote the smallest and largest eigenvalues of $X$ as
\begin{align}
	\lambda_{min}(X) = \inf_{ v \in \mathbb{R}^{n} ,  ||v||_{2}=1} v^{T} X v ,  \quad 	\lambda_{max}(X) = \sup_{v \in \mathbb{R}^{n} , ||v||_{2}=1} v^{T} X v
\end{align}
For any real symmetric matrix $X$, the set of eigenvalues of $X$ will be denoted as $\text{spec}(X) $. The spectrum of a tensor product of two matrices is the tensor product of each spectra, 
\begin{align*}
\text{spec}(  X \otimes Y ) = \text{spec}(  X  ) \otimes \text{spec}(  Y  )
\end{align*} 

For any two matrices $X$ and $Y$ of the same size, the Frobenius inner product $\langle \cdot , \cdot  \rangle_{F} $ is defined as the quantity
\begin{align}
	X ,Y \in \mathbb{R}^{n \times k} , \quad	\langle X , Y \rangle_{F} = \text{Tr}[ X^{T} Y ] = \sum_{i=1}^{n} \sum_{j=1}^{k} X_{ij} Y_{ij}
\end{align}

The Frobenius norm $|| \cdot ||_{F} = [ \langle \cdot , \cdot  \rangle_{F}]^{\frac{1}{2}}$ of a matrix is defined as
\begin{align}
	& X \in \mathbb{R}^{n \times k} , \quad || X ||_{F} = ( \text{Tr}[X^{T}X] )^{\frac{1}{2}} = (\sum_{i=1}^{n} \sum_{j=1}^{k} |X_{ij}|^{2} )^{\frac{1}{2}}
\end{align}
The Frobenius inner product is stronger than sub-multiplicative, 
\begin{align}\label{F Inner Product}
	\langle X , Y \rangle_{F} \leq ||X||_{2} ||Y||_{F} \leq ||X||_{F} ||Y||_{F}
\end{align}
where $||X||_{2} ||Y||_{F}$ can significantly smaller than $||X||_{F} ||Y||_{F}$.

It is often more convenient to work with the Frobenius norm then the matrix 2-norm. They are related by the simple inequality,
\begin{align}
	\frac{1}{\sqrt{r}}||X||_{F} \leq || X ||_{2} \leq || X ||_{F}
\end{align}
where $r$ is the rank of $X$. 

A square matrix $A$ is positive semi-definite if for all $v \in \mathbb{R}^{n}$ we have that $ v^{T} A v  \geq 0$. Positive definite matrices are denoted as $0 \preceq A$. $ A \preceq B$ if and only if $0 \preceq B - A$.
Both the spectral norm and the Frobenius norm are monotone with respect to positive definite ordering meaning that if $X \preceq Y$, then $|| X ||_{F} \leq ||Y||_{F}$ and  $|| X ||_{2} \leq ||Y||_{2}$.

\section{Group and Representation Theory}

We establish some notation and review some elements of group theory and representation theory. For a comprehensive review of representation theory, please see \cite{Zee_2016_Group}. The identity element of any group $G$ will be denoted as $e$. We will always work over the field $\mathbb{R}$ unless otherwise specified.

\subsection{Group Theory}

A group $G$ is a non-empty set combined with a associative binary operation $\cdot : G \times G \rightarrow G$ that satisfies the following properties
\begin{align*}
	& \text{existence of identity: } e \in G, \text{ s.t. } \forall g\in G, \enspace e \cdot g = g \cdot e = g  \\
	& \text{existence of inverse: } \forall g \in G, \exists g^{-1} \in G, \enspace g \cdot g^{-1} = g^{-1} \cdot g = e  \\
\end{align*}

\subsection{Representation Theory}
Let $V$ be a vector space over the field $\mathbb{C}$. A representation $(\rho , V)$ of a group $G$ consists of $V$ and a group homomorphism $\rho : G \rightarrow \text{Hom}[V,V] $. By definition, the $\rho$ map satisfies
\begin{align*}
	\forall g,g' \in G, \enspace \forall v \in V, \enspace \rho(g)\rho(g')v = \rho(gg') v 
\end{align*}
Heuristically, a group can be thought of as the embedding of an group (which is an abstract mathematical object) into a set of matrices. Two representation $(\rho,V)$ and $(\sigma,W)$ are said to be equivalent representations if there exists a unitary matrix $U$
\begin{align*}
	\forall g\in G, \quad U\rho(g) = \sigma(g)U
\end{align*}
The linear map $U$ is said to be a $G$-intertwiner of the $(\rho,V)$ and $(\sigma , W)$ representations. The space of all $G$-intertwiners is denoted as $\Hom_{G}[ (\rho,V) , (\sigma , W) ]$. Specifically,
\begin{align*}
\Hom_{G}[ (\rho,V) , (\sigma , W) ] = \{ \Phi : V \rightarrow W | \enspace \forall g\in G, \enspace \Phi \rho(g) = \sigma(g) \Phi , \enspace \Phi \text{ is linear} \}
\end{align*}
Note that $\Hom_{G}[ (\rho,V) , (\sigma , W) ]$ is a vector space. Much of classical group theory studies the structure of the intertwiners of representations \cite{Ceccherini_2008_Harmonic}.
The unitary theorem in representation theory \cite{Ceccherini_2008_Harmonic} says that all representations of compact groups are equivalent to a unitary representation. A representation is said to be reducible if it breaks into a direct sum of smaller representations. Specifically, a unitary representation $\rho$ is reducible if there exists an unitary matrix $U$ such that
\begin{align*}
\forall g\in G, \quad \rho(g) = U [ \bigoplus_{i=1}^{k} \sigma_{i}(g) ] U^{\dagger}
\end{align*}
where $k\geq2$ and $\sigma_{i}$ are smaller irreducible representations of $G$. The set of all non-equivalent representations of a group $G$ will be denoted as $\hat{G}$. All representations of compact groups $G$ can be decomposed into direct sums of irreducible representations. Specifically, if $(\sigma , V)$ is a $G$-representation, 
\begin{align*}
(\sigma , V) = U[  \bigoplus_{ \rho \in \hat{G} } m^{\rho}_{\sigma}(\rho , V_{\rho} )  ] U^{\dagger}
\end{align*}
where $U$ is a unitary matrix and the integers $m^{\rho}_{\sigma}$ denote the number of copies of the irreducible $(\rho , V_{\rho} )$ in the representation $(\sigma , V)$.

\subsection{Schur's Lemma}\label{Suppl_Schur_Lemma}

Schur's lemma is one of the fundamental results in representation theory \cite{Zee_2016_Group}. Let $G$ be a compact group. Let $(\rho , V)$ and $(\sigma , W)$ be irreducible representations of $G$. Then, 
Schur's lemma states the following: Let $\Phi: V \rightarrow W$ be an intertwiner of  
$(\rho , V)$ and $(\sigma , W)$. Then, $\Phi$ is either zero or the proportional to the identity map. In other words,
\begin{align*}
	\text{ if } \forall g\in G, \enspace	\Phi  \rho(g) = \sigma(g) \Phi \implies \begin{cases}
		& \Phi \propto \mathbb{I} \text{ if }(\rho , V) = (\sigma , W)  \\
		& \Phi = 0 \text{ if else}
	\end{cases}
\end{align*}
Equivalently, if $(\rho , V)$ and $(\sigma , W)$ are irreducible representations, the space of intertwiners of representations satisfies
\begin{align*}
	\Hom_{G}[ (\rho , V), (\sigma , W)  ] \cong \begin{cases}
		&\mathbb{C} \text{ if } (\rho , V) = (\sigma , W)  \\
		& 0 \text{ if else}
	\end{cases}
\end{align*}
A corollary of Schur's lemma is the following: Let $(\rho , V)$ be a irreducible representation of $G$. Let $M \in \mathbb{C}^{d_{\rho}\times d_{\rho}}$ be a matrix. Suppose that 
\begin{align*}
	\forall g \in G, \quad \rho(g)M = M\rho(g)
\end{align*}
holds. Then, $M$ is proportional to the identity matrix. The constant of proportionally can be determined by taking traces. Specifically,
\begin{align*}
    M = \frac{\text{Tr}[M]}{d_{\rho}} \mathbb{I}_{d_{\rho}}
\end{align*}

\paragraph{Extended Shur Lemma}\label{Suppl_Extended_Schur_Lemma}

Schur's Lemma can be extended to reducible representations. Let $(\rho,V_{\rho})$ and $(\sigma, W_{\rho})$ be $G$ representations which decompose into irriducibles as
\begin{align*}
(\rho,V_{\rho}) = U[\bigoplus_{ \tau \in \hat{G} } m^{\rho}_{\tau} (\tau , W_{\tau}   ) ]U^{\dagger} \quad (\sigma,V_{\sigma}) = V[ \bigoplus_{ \tau \in \hat{G} } m^{\sigma}_{\tau} (\tau , W_{\tau}   ) ] V^{\dagger}
\end{align*}
where $U, V$ are fixed unitary matrices that diogonlize the $\rho$ and $\sigma$ representations, respectively. Then, the vector space of intertwiners between $(\rho,V_{\rho})$ and $(\sigma, W_{\sigma})$ has dimension
\begin{align*}
\dim \Hom_{G}[ (\rho,V_{\rho}) , (\sigma,V_{\sigma}) ] = \sum_{ \tau \in \hat{G} }  m^{\rho}_{\tau}m^{\sigma}_{\tau}
\end{align*}
Furthermore, elements of the space $\Hom_{G}[ (\rho,V_{\rho}) , (\sigma,V_{\sigma}) ]$ have block structure. Specifically, any $\Phi \in \Hom_{G}[ (\rho,V_{\rho}) , (\sigma,V_{\sigma}) ]$ can be parameterized in block diagonal form as
\begin{align*}
\Phi = U[ \bigoplus_{ \tau \in \hat{G} } \Phi^{\tau} ]V^{\dagger}
\end{align*}
and each block $\Phi^{\tau}$ can be written as
\begin{align*}
\Phi^{\tau} = \begin{bmatrix}
\Phi^{\tau}_{11} \mathbb{I}_{ d_{\tau}  } & \Phi^{\tau}_{12} \mathbb{I}_{ d_{\tau}  } & ... & \Phi^{\tau}_{1m^{\sigma}_{\tau}} \mathbb{I}_{ d_{\tau}  } \\
\Phi^{\tau}_{21} \mathbb{I}_{ d_{\tau}  } & \Phi^{\tau}_{22} \mathbb{I}_{ d_{\tau}  } & ... & \Phi^{\tau}_{2m^{\sigma}_{\tau}} \mathbb{I}_{ d_{\tau}  } \\
... & ... & ... & ... \\
\Phi^{\tau}_{m^{\rho}_{\tau}1} \mathbb{I}_{ d_{\tau}  } & \Phi^{\tau}_{m^{\rho}_{\tau}2} \mathbb{I}_{ d_{\tau}  } & ... & \Phi^{\tau}_{m^{\rho}_{\tau}m^{\sigma}_{\tau}} \mathbb{I}_{ d_{\tau}  } \\
\end{bmatrix}
\end{align*}
where each $\Phi^{\tau}_{ij} \in \mathbb{C}$ is a complex constant and $\mathbb{I}_{d_{\tau}} $ is the identity matrix of the $\tau$ irreducible dimension $d_{\tau} = \dim (\tau , W_{\tau} )$.

\subsection{Irreducible Representation Orthogonality Relations}

Matrix elements of irriducibles representations satisfy a set of orthogonality relations \cite{Zee_2016_Group}. Specifically, let $\rho$ and $\sigma$ be irreducible representations of the group $G$. Then,
\begin{align*}
\sum_{g \in G } \rho_{kk'}(g)\sigma(g)_{nn'}^{\dagger} = \frac{|G|}{d_{\rho}} \delta_{\rho,\sigma} \delta_{kn} \delta_{k'n'}
\end{align*}
where $|G|$ is the cardinality of the group. Let $G$ be a compact group. The character of a representation is a map $\chi_{\rho} : G \rightarrow \mathbb{C}$ defined as
\begin{align*}
\chi_{\rho}(g) = \text{Tr}[  \rho(g)  ]
\end{align*}
The character is invariant under normal transformations $\chi_{\rho}(g) = \chi_{\rho}(hgh^{-1})$. Furthermore, the character is independent of the choice of basis of the representation. Specifically, under a change of basis $\rho(g) \rightarrow U \rho(g) U^{\dagger}$, the character is unchanged as $\chi_{U\rho U^{\dagger}}(g) = \text{Tr}[ U \rho(g) U^{\dagger} ] = \text{Tr}[U \rho(g) U^{\dagger}] = \chi_{\rho}(g)$. Any two irriducibles $\rho$ and $\rho'$ characters satisfy the orthogonality relation, 
\begin{align*}
\int_{ g \in G}dg \text{ }\chi_{\rho}(g)\chi_{\rho'}(g) = \delta_{\rho \rho'} |G|
\end{align*}
Furthermore, characters of groups form a complete basis over the set of normal complex valued functions defined on $G$ \cite{Ceccherini_2008_Harmonic}.

\subsection{Peter-Weyl Theorem}\label{Suppl_Peter-Weyl}
The Peter-Weyl theorem \cite{Ceccherini_2008_Harmonic} states that all representations of compact groups can be decomposed into a countably infinite sets of irreducible representations. Consider the set of functions
\begin{align*}
\mathcal{F} = \{  \enspace f \enspace | \enspace f: G \rightarrow \mathbb{C} \enspace \}
\end{align*}
of all complex valued function defined on $G$. The set $\mathcal{F}$ forms a vector space over the field $\mathbb{C}$. The group $G$ acts on vector space $\mathcal{F}$. Specifically, define the group action $\lambda: G \times \mathcal{F} \rightarrow \mathcal{F}$ as
\begin{align*}
\forall f\in \mathcal{F},\enspace \forall g,g'\in G, \quad (\lambda_{g} \cdot  f)(g') = f(g^{-1}g) \in \mathcal{F}
\end{align*}
The action satisfies $\lambda_{g}\lambda_{g'}=\lambda_{gg'}$ and is a group homeomorphism. The left-regular representation of a group is defined as $ (\lambda, \mathcal{F})$. The Peter-Weyl theorem \cite{Ceccherini_2008_Harmonic} states that
\begin{align*}
(\lambda, \mathcal{F}) = U[ \bigoplus_{\rho \in \hat{G}} d_{\rho}(\rho , V_{\rho} )  ] U^{\dagger}
\end{align*}
where $U$ is the unitary matrix. Thus, the left-regular representation decomposes into $d_{\rho}$ copies of each $(\rho , V_{\rho})$ irreducible. In other words, the Peter-Weyl theorem states that matrix elements of irreducible $G$-representations form an orthonormal base of the space of square integrable functions on $G$.

\subsubsection{Fourier Transform on Groups}\label{Fourier Transform on Groups}
The Peter-Weyl Theorem allows for the Fourier Transform on groups \cite{Ceccherini_2008_Harmonic}. Specifically, let $G$ be a compact group. Let $f \in \mathcal{F}$. The Fourier coefficient with respect to the $\rho$-th irreducible is defined as
\begin{align*}
\hat{f}^{\rho} = \int_{g\in G}dg\text{ }f(g)\rho(g)
\end{align*}
where each $\hat{f}^{\rho} \in \mathbb{C}^{d_{\rho} \times d_{\rho} } $ is a complex $d_{\rho} \times d_{\rho}$ matrix. The Inverse Fourier transform is then defined as
\begin{align*}
f(g) = \sum_{ \rho \in \hat{G} } d_{\rho}\text{Tr}[ \rho(g^{-1}) \hat{f}^{\rho} ]
\end{align*}
\paragraph{Delta Function Fourier Transform}\label{Delta Function Fourier Transform on Groups}
Consider the delta function centered at $g' \in G$,
\begin{align}
\delta_{ g' }(g) = \begin{cases}
   &  1 \text{ if } g=g' \\
    & 0 \text{ else}
\end{cases}
\end{align}
The Fourier transform is easy to compute
\begin{align*}
\hat{\delta}^{\rho}_{ g' } = \int_{ g\in G }dg\text{ } \delta_{g'}(g)\rho(g) = \rho(g')
\end{align*}
Thus, for the delta function $\hat{\delta}^{\rho}_{g'} = \rho(g')$. Using the inverse Fourier transformation, we have that
\begin{align*}
\delta_{g'}(g) = \sum_{ \rho \in \hat{G} } d_{\rho}\text{Tr}[ \rho(g^{-1}) \rho(g')  ]
\end{align*}

\paragraph{Fourier Transform on Abelian Groups}
The standard Fourier Transform can be recovered by considering abelian groups. Specifically, let $\mathbb{Z}_{N}$ be the cyclic group of order $N$. Let $g$ be a generator of $\mathbb{Z}_{N}$. Then, all irreducible representations of $\mathbb{Z}_{N}$ are one dimensional and take the form
\begin{align*}
\rho_{k}( g^{n} ) = \exp(\frac{ikn}{N})
\end{align*}
where $1 \leq k \leq N$ is an integer. Then, using the Fourier Transform on groups decomposition \ref{Fourier Transform on Groups}, we have that
\begin{align*}
& \hat{f}_{k} = \sum_{i=1}^{N} f_{n} \exp(\frac{ikn}{N}) \quad f_{n} = \frac{1}{N} \sum_{i=1}^{N} \hat{f}_{k} \exp( - \frac{ikn}{N}) 
\end{align*}
which is the standard Fourier transformation for a discrete single-variable waveform.

\subsection{Parseval$-$Plancherel Theorem on Compact Groups}\label{Suppl_Parseval-Plancherel}

The Parseval–Plancherel theorem relates the $L^{2}(G)$ norm of a function to the norm of its Fourier coefficients. Specifically, let $f: G \rightarrow \mathbb{C}$. Consider the Fourier expansion of $f$,
\begin{align*}
f(g) = \sum_{ \rho \in \hat{G} } \sum_{kk'=1}^{d_{\rho} } d_{\rho}f^{\rho}_{kk'}\rho_{kk'}(g)
\end{align*}
Now, consider the $L^{2}(G)$ norm of $f$,
\begin{align*}
|| f ||_{ L^{2}(G) }^{2} = \int_{g \in G}dg\text{ } |f(g)|^{2}
\end{align*}
Using the Fourier expansion, we have that
\begin{align*}
|| f ||_{ L^{2}(G) }^{2} = \int_{g \in G}dg\text{ } |f(g)|^{2} = \int_{g \in G}dg\text{ } \sum_{ \rho \rho' \in \hat{G} } d_{\rho} d_{\rho'}f^{\rho}_{kk'}\rho_{kk'}(g) (f^{\rho}_{kk'})^{\dagger} \rho_{kk'}(g)^{\dagger}
\end{align*}
Using the orthogonality relations, we have that
\begin{align*}
\int_{g \in G}dg\text{ } \sum_{ \rho \rho' \in \hat{G} } d_{\rho} d_{\rho'}f^{\rho}_{kk'}\rho_{kk'}(g) (f^{\rho}_{kk'})^{\dagger} \rho_{kk'}(g)^{\dagger} = \sum_{ \rho \in \hat{G} } d_{\rho}^{2} || \hat{f}^{\rho} ||^{2}_{F}
\end{align*}
Ergo,
\begin{align*}
|| f ||_{L^{2}(G)}^{2} = \sum_{\rho \in \hat{G} } d_{\rho}^{2} || \hat{f}^{\rho} ||^{2}_{F}
\end{align*}
This is known as the Parseval–Plancherel relation on compact groups and relates the function space norm of $f$ to the Fourier transformation of $f$.

\section{Irreducible Representations of $SO(2)$}

\begin{figure}[h]\label{Table:SO2_Fourier_Transform}
\centering
\begin{tabular}{|l|}
\hline
\includegraphics[width=0.45\textwidth]{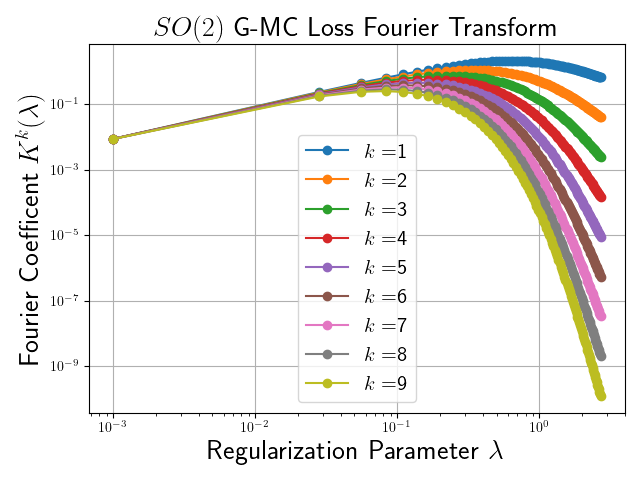} 
\includegraphics[width=0.45\textwidth]{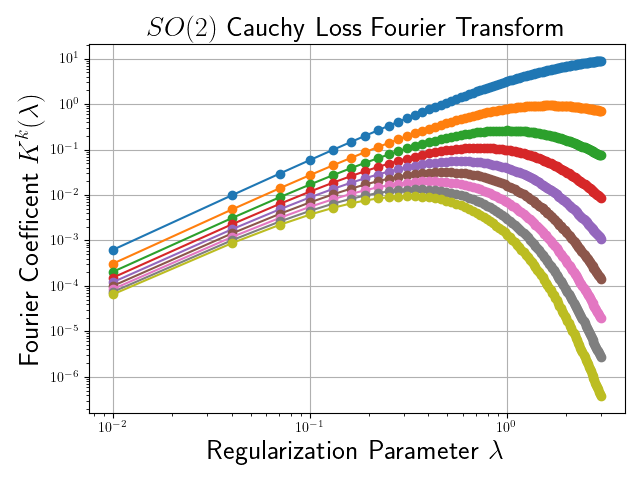} \\
\hline
\end{tabular}
\vspace*{-0.25cm}
\caption{ $SO(2)$ Fourier Transform coefficient $K^{k}(\lambda)$ for different irreducible representation $k$ of $SO(2)$. Both the G-MC loss and Cauchy loss have positive definite Fourier coefficient $K^{k}(\lambda)$ for all $\lambda$ and all $k$. }
\end{figure}

Complex irreducible representations of $SO(2)$ are labeled by an integer $k$ and have all dimension $d_{k} = 1$. They are parameterized by
\begin{align*}
	\forall \phi \in SO(2), \quad \sigma_{k}(\phi) = \exp(ik\phi)
\end{align*}
Real irreducible representations of $SO(2)$ are labeled by an non-negative integer $k$ and have dimension $d_{0} = 1$ and $d_{k\geq1} = 2$. Real $SO(2)$ irreducible representations are parameterized by,
\begin{align*}
	\rho_{0}(\phi) = \sigma_{0}(\phi) = 1 
\end{align*}
and for $k\geq1$,
\begin{align*}
	\rho_{k}(\phi) = \sigma_{k}(\phi) \oplus \sigma_{-k}(\phi)  = \begin{bmatrix}
		\cos(k \phi) & \sin(k \phi) \\
		-\sin(k \phi) & \cos(k \phi) \\
	\end{bmatrix}
\end{align*}

\paragraph{Fourier Transform over $SO(2)$}
The complex Fourier Transform on $SO(2)$ is the standard grade school Fourier Transformation. Specifically, let $f : S^{1} \rightarrow \mathbb{C}$ be a complex valued $2\pi$-periodic function. Any $f$ can be decomposed as
\begin{align*}
	f(\phi) = \sum_{k=-\infty}^{\infty} \hat{f}^{c}_{k} \exp(ik\phi)
\end{align*}
where the complex valued Fourier coefficients are computed via
\begin{align*}
\hat{f}^{c}_{k} = \frac{1}{2\pi} \int_{0}^{2\pi}d\phi \text{ }f(\phi) \exp(-ik\phi)
\end{align*}
The real Fourier Transform on $SO(2)$ is slightly more complicated. Let $f : SO(2) \rightarrow \mathbb{R}$ be a real function defined on $SO(2)$. Then, Fourier coefficients $f_{k}$ are given by
\begin{align*}
\hat{f}_{0} = \frac{1}{2\pi} \int_{0}^{2\pi}d\phi \text{ }f(\phi)
\end{align*}
and for $k\geq 1$,
\begin{align*}
\hat{f}_{k} = \frac{1}{2\pi} \int_{0}^{2\pi}d\phi \text{ }f(\phi) \begin{bmatrix}
    \cos(k\phi) & \sin(k\phi) \\
    -\sin(k\phi) & \cos(k\phi) \\
\end{bmatrix}
\end{align*}
The real valued Fourier coefficients $f_{k}$ can be related to the complex Fourier coefficients $f^{c}_{k}$ by
\begin{align*}
f_{0} = f_{0}^{c} , \quad f_{k} =  f^{c}_{k} + \Bar{f}^{c}_{k}
\end{align*}

\paragraph{ $F$-Norm Loss }
Now, consider the quadratic $F$-norm loss centered at point $\phi'$, \begin{align*}
	f_{\phi'}(\phi) = || \rho_{1}(\phi) - \rho_{1}(\phi') ||_{F}^{2} = 4[ 1 - \cos( \phi - \phi') ] 
\end{align*}
The Fourier transform has elements given by
\begin{align*}
	\hat{f}^{k}_{\phi'} = \frac{1}{2\pi}\int_{0}^{2\pi} d\phi \text{ } 4[1 - \cos( \phi - \phi') ]  \exp(-ik\phi) 
\end{align*}
Thus, with respect to the complex irreducible $\sigma_{k}$, we have that
\begin{align*}
\hat{f}^{k}_{\phi'} = 2[ 2\delta_{k} - \delta_{k-1} - \delta_{k+1} ] \exp(-ik\phi)
\end{align*}

\section{Irreducible Representations of $SO(3)$}

Irreducible representations of $SO(3)$ are labeled by an integer $\ell$ and have dimension $d_{\ell} = 2\ell + 1$. Following convention, the $SO(3)$ representations are specified using the Wigner $D$ matrices $D^{\ell}$. Using Bra-Ket notation \cite{Landau_1991_Quantum}, $SO(3)$ representations are defined as
\begin{align*}
	D^{\ell}_{kk'}(R) = \langle \ell k | R | \ell k' \rangle
\end{align*}
where $|\ell k \rangle$ are states of total angular momentum $\ell$. The Wigner $D$-matrices are more related to the more familiar spherical harmonics $Y^{k}_{\ell}( \hat{n}) = \langle \hat{n} | \ell k \rangle $ \cite{Landau_1991_Quantum}. Specifically, under a rotation
\begin{align*}
	Y^{k}_{\ell}( R^{-1} \hat{n} ) = \sum_{k'=-\ell}^{\ell} D^{\ell}_{kk'}(R) Y^{k'}_{\ell}( \hat{n} )
\end{align*}
so that under rotation the vector $Y^{\ell}$ transforms like the irreducible $\ell$-representation.

\paragraph{Fourier Transform over $SO(3)$}

Let $f: SO(3) \rightarrow \mathbb{C}$. The Fourier coefficients $\hat{f}^{\ell}$ over $SO(3)$ are then given by
\begin{align*}
	\hat{f}^{\ell} = \int_{ g \in SO(3) } dg \text{ } D^{\ell}(g) f(g) 
\end{align*}
where $dg$ is the Haar-measure over $SO(3)$. 

\paragraph{ $F$-Norm Loss}\label{Suppl_F_Norm_Loss}
Similarly, consider the quadratic loss
\begin{align*}
f_{g'}(g) = || D^{1}(g) - D^{1}(g') ||^{2}_{F}
\end{align*}
Using the result of \ref{Suppl_Thrm_I}, we have that
\begin{align*}
\hat{f}^{\ell}_{g'} = K^{\ell}D^{\ell}(g')
\end{align*}
where
\begin{align*}
K^{\ell} = \frac{1}{2\ell + 1 } \int_{ g \in SO(3) } dg \text{ }  || D^{1}(g) - \mathbb{I}_{3} ||^{2}_{F} \chi^{1}(g)
\end{align*}
Using the algebraic identity
\begin{align*}
|| D^{1}(g) - \mathbb{I}_{3} ||^{2}_{F} = 6 - 2\text{Tr}[ D^{1}(g) ] = 6-2\chi^{1}(g)
\end{align*}
we have that
\begin{align*}
K^{\ell} = \frac{2}{2\ell + 1 }\int_{ g \in SO(3) } dg \text{ } \chi^{\ell}(g) [ 3 - \chi^{1}(g) ]
\end{align*}
Now, using the identities
\begin{align*}
\int_{ g \in SO(3) } dg \text{ } D^{\ell}(g) = \delta_{0\ell},  \quad  \int_{ g \in SO(3) } dg \text{ } \chi^{\ell}(g) \chi^{1}(g)  = \int_{ g \in SO(3) } dg \text{ } \chi^{\ell \otimes 1}(g) 
\end{align*}
Now, note that $\chi^{\ell}(g)\chi^{1}(g)$ is the tensor product representation $1 \otimes \ell =  (\ell - 1) \oplus \ell \oplus (\ell + 1) $ representation. Thus,
\begin{align*}
K^{\ell} = \frac{2}{2\ell + 1 }[ 3\delta_{\ell} - \delta_{\ell-1}  \enspace ]
\end{align*}
Thus, for the quadratic $F$-norm loss, the only non-zero Fourier Transform coefficients are the $\ell=0$ and $\ell=1$ irriducibles. Thus, when using quadratic $F$-norm loss, our proposed algorithm reduces to \cite{Doherty_2022}.

\begin{figure}[h]
\centering
\begin{tabular}{|l|}
\hline
\includegraphics[width=0.49\textwidth]{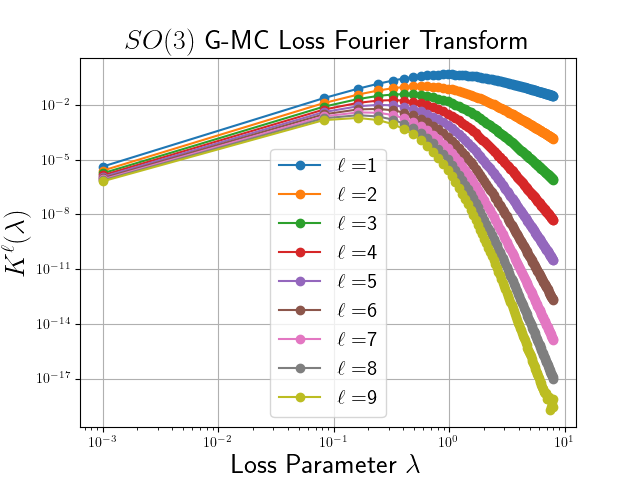}
\includegraphics[width=0.49\textwidth]{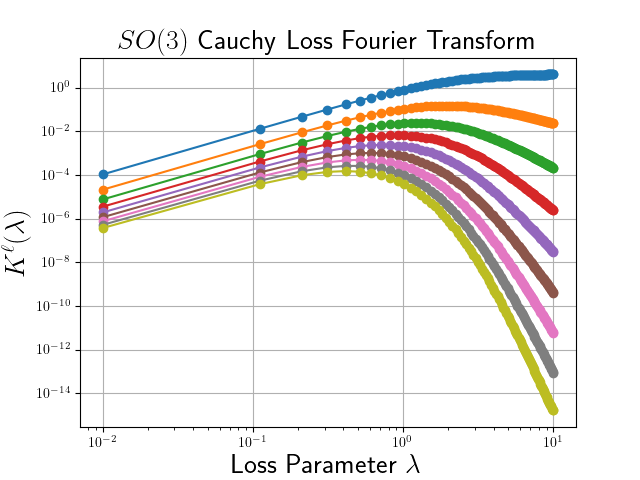} \\
\hline
\end{tabular}
\vspace*{-0.25cm}
\caption{ $SO(3)$ Fourier Transform coefficient $K^{\ell}(\lambda)$ for different irreducible representation $\ell$ of Robust Loss functions as a function of regularization parameter $\lambda$. Left:German-McClure (G-MC) loss. Right: Cauchy loss. Both the G-MC loss and Cauchy loss have positive definite Fourier coefficient $K^{\ell}(\lambda)$ for all $\lambda$ and all $\ell$.  }\label{Table:SO3_Fourier_Transform_Losses}
\end{figure}

\section{Langevin Probability Density}\label{Suppl_Lang}
The Langevin probability density \eqref{Langevin Probility Density} is invariant under normal transformations $R \rightarrow ORO^{T}$ with $O \in SO(d)$ as	
$\text{Pr}[ O R O^{T} ] = \text{Pr}[R]$ holds. This means that $SO(d)$ is invariant under class transformation and we can thus use the Weyl character formula to evaluate $c_{d}(k)$ \cite{Boumal_2013, Weyl_1926}. We are especially interested in the cases $d=2,3$. We have that,
\begin{align}
	& \nonumber c_{2}(k) = \frac{1}{2\pi} \int_{R\in SO(2)} dR \exp(k \text{Tr}[R])   = \frac{1}{2\pi} \int_{-\pi}^{\pi} \exp( 2k \cos(\phi) ) d \phi = I_{0}(2k)
\end{align}
and 
\begin{align}
	& \nonumber c_{3}(k) = \frac{1}{2\pi} \int_{R\in SO(3)} dR \exp(k \text{Tr}[R])  =  \frac{1}{2\pi} \int_{-\pi}^{\pi} (1 - \cos(\phi)) \exp( k (1 + 2\cos(\phi) ) ) d \phi \\
	& \nonumber = \exp(k) (I_{0}(2k) - I_{1}(2k))
\end{align}
To summarize, the normalization for the isotropic Langevin distribution is given by

\begin{align}\label{Normilization of Isotropic Langevin Distribution}
	c_{d}(k) = \begin{cases}
		I_{0}(2k), & d = 2, \\
		\exp(k) ( I_{0}(2k) - I_{1}(2k) ), & d=3.
	\end{cases}
\end{align}

\subsection{Fitting Loss Functions to Langevin Noise Models}\label{Suppl_Lang_Fit}

Robust methods require an nominal model to capture systematic noise. Outliers are determined by measuring the deviation of observed data from the systemic noise model. In this work, we model systemic noise using a Langevin model (this is a natural assumption because the Langevin distribution is the maximum entropy distribution on an $SO(d)$ random variable) and we use a uniform noise corruption model as a stand in for epistemic noise. Consider the deviation of a Langevin random variable from the identity matrix in $d$ dimensions, $|| R - \mathbb{I}_{d} ||_{F}$. Using the results of \cite{Rosen_2019}, for large concentration parameter $k$ the quantity $ k || R - \mathbb{I}_{d} ||^{2}_{F}$ behaves like a $\chi^{2}$-random variable with $d$ degrees of freedom,
\begin{align*}
k || R - \mathbb{I}_{d} ||^{2}_{F} \sim \chi^{2}_{d}
\end{align*}
Thus, using the Laurent-Massart bound (cite), for large $k$, the probability that $ k || R - \mathbb{I}_{d} ||^{2}_{F}$ is roughly three standard deviations from zero is
\begin{align*}
\text{Pr}[ k || R - \mathbb{I}_{d} ||^{2}_{F} \geq d + 2\epsilon + 2\sqrt{\epsilon d}  ] \leq \exp(-\epsilon)
\end{align*}
Inserting $\epsilon = 5 $ gives
\begin{align*}
\text{Pr}[ k || R - \mathbb{I}_{d} ||^{2}_{F} \geq d + 10 + 2\sqrt{10d}  ] \leq \exp(-5) \approx 0.01
\end{align*}
so that the probability that $|| R - \mathbb{I}_{d} ||_{F} \leq \frac{d + \sqrt{10}}{\sqrt{k}}$ is greater than 99 percent. We thus choose the robust loss function to deviate from the ordinary least squares solution at roughly $x = \frac{d + \sqrt{10}}{\sqrt{k}}$.

\begin{figure}[h]\label{Table:Loss_functions}
\centering
\begin{tabular}{|l|}
\hline
\includegraphics[width=0.45\textwidth]{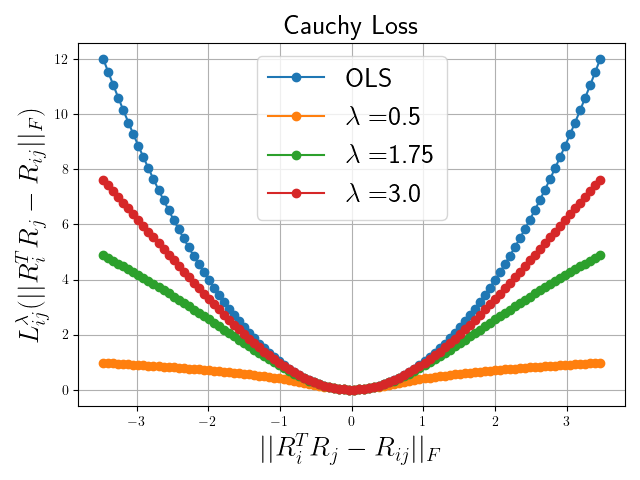} 
\includegraphics[width=0.45\textwidth]{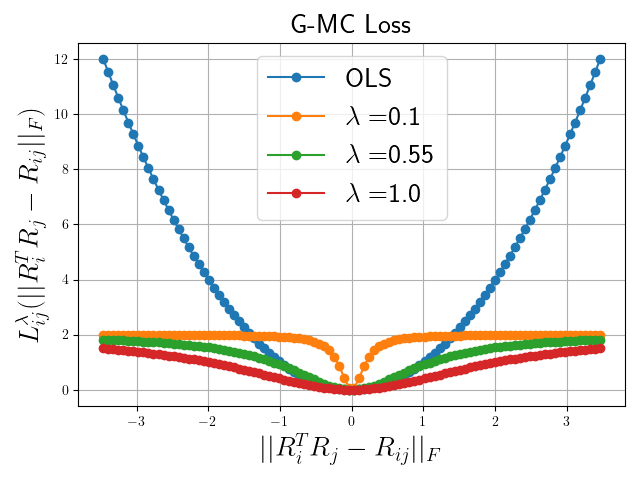} \\
\hline
\end{tabular}
\vspace*{-0.25cm}
\caption{ Left: Cauchy loss function $L^{\lambda}$ for different regularization parameters $\lambda$. Right: GM-C loss function $L^{\lambda}$ for different regularization parameters $\lambda$. The robust loss functions plateau far away from zero. The parameter $\lambda$ determines the sensitivity of the loss function to outlier rejection. For an in-depth discussion of how we choose $\lambda$, please see \ref{Suppl_Lang_Fit}.    }
\end{figure}

\subsection{Germane McClure (G-MC) Loss}
Consider the German McClure (G-MC) Loss $L^{\lambda}(x) = \frac{2x^{2}}{x^{2} + 4\lambda^{2}}$. The second derivative of the G-MC loss,
\begin{align*}
\frac{d^{2}}{dx^{2}} L^{\lambda}(x) = 2\lambda^{2} \frac{ 4\lambda^{2} - 3x^{2} }{ ( 4\lambda^{2} + x^{2} )^{3} }
\end{align*}
with inflection points at $x = \pm \frac{2|\lambda|}{\sqrt{3}}   $.

We then impose the constraint that the G-MC loss inflection points are at $x = \frac{d + \sqrt{10}}{\sqrt{k}}$. Solving for $\lambda$ gives the G-MC policy $\lambda = \lambda(k)$ regularization policy as
\begin{align*}
\lambda^{\text{G-MC}}(k) = \frac{d\sqrt{3} + \sqrt{30}}{ 2\sqrt{k}}
\end{align*}
Note that for large $k$, $\lambda^{\text{G-MC}}(k)$ tends to zero, illustrating that for highly concentrated Langevin variables, any deviation from the identity matrix is an outlier. Thus, when using the G-MC edge loss for an edge with parameter $k_{ij}$, the $ij$-th edge regularization parameter $\lambda^{\text{G-MC}}_{ij} = \frac{d\sqrt{3} + \sqrt{30}}{ 2\sqrt{k_{ij}}} $.

\subsection{ Cauchy Loss }
Now, consider the Cauchy loss $L^{\lambda}(x) = \lambda^{2} \log( 1 + \frac{x^{2}}{\lambda^{2}} )$. The second derivative of the Cauchy loss is given by
\begin{align*}
\frac{d^{2}}{dx^{2}} L^{\lambda}(x)  = \frac{2\lambda^{2}( \lambda^{2} - x^{2} )}{ (1+x^{2})^{2} }
\end{align*}
with inflection points at $x = \pm | \lambda |$.  We then impose the constraint that the Cauchy loss function inflection points are at $x=\frac{d + \sqrt{10}}{\sqrt{k}}$. Solving for $\lambda$ gives the G-MC policy $\lambda = \lambda(k)$ regularization policy as
\begin{align*}
\lambda^{\text{cauchy}}(k) = \frac{d + \sqrt{10}}{\sqrt{k}}
\end{align*}
Note that for large $k$, $\lambda^{\text{cauchy}}(k)$ tends to zero, illustrating that for highly concentrated Langevin variables, any deviation from the identity matrix is an outlier.
Similar choices of regularization parameters as a function of edge weights can be derived for other loss functions.

\section{Numerical Experiments}

\begin{figure}[h!]\label{Table:small_world_graphs}
	\begin{tabular}{|l|}
		\hline
		\includegraphics[width=0.45\textwidth]{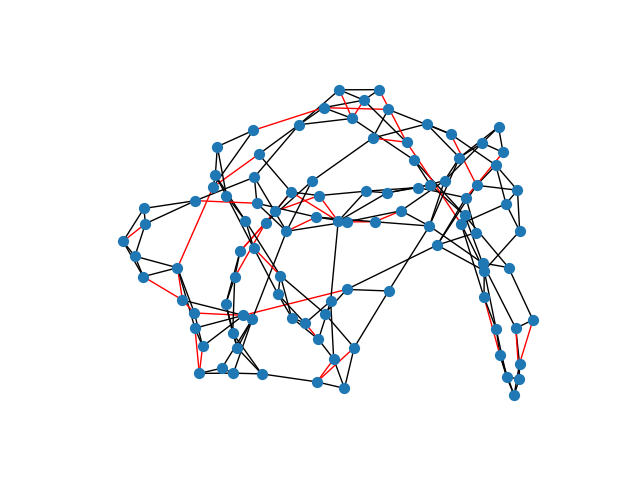} 
		\includegraphics[width=0.45\textwidth]{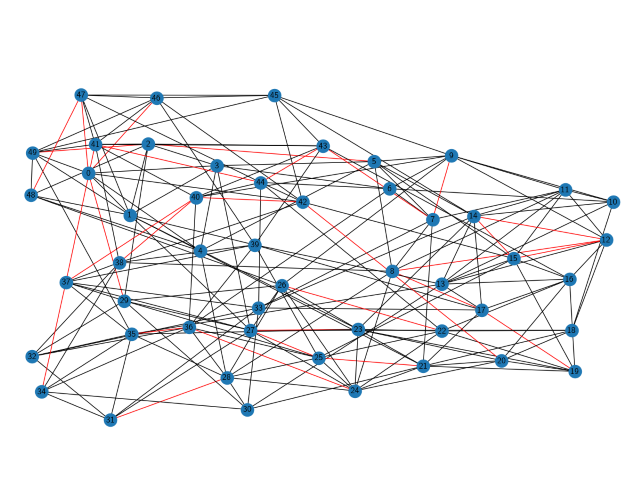} \\
		\hline
	\end{tabular}
	\vspace*{-0.25cm}
	\caption{ Left: The graph generated by NetworkX with 100 nodes and local connectivity parameters $k=6$, used in $SO(2)$ tests. Right: The graph generated by NetworkX with 50 nodes and local connectivity $k=8$, used in $SO(3)$ tests. The edges depicted in black represent those where noise is introduced via the Langevin distribution, while the edges in red gets completely random noise. The rewire probability was set to be $0.3$. }
	\label{Small world}
\end{figure}

\subsection{Numerical Experiments on $SO(2)$ Rotation Averaging}

We test our results on two dimensional rotation averaging problems. Graphs are constructed utilizing the Python package NetworkX, with rotations assigned to each edge. Noise is then added to each edge, either drawn from a Langevin distribution or a uniform distribution. Figure \ref{Small world} shows the graph generated by NetworkX. Table \ref{fig:SO2_edge_fail_comparison_II}, \ref{fig:SO2_edge_fail_comparison_III}, \ref{fig:SO2_edge_fail_comparison_IV} presents a comparative analysis of the efficacy of our methodology relative to varying quantities of irreducibles used in graphs, which are formulated using the NetworkX library with different local connectivity parameters.

\begin{figure*}[h!]
\begin{subtable}{0.80\paperwidth}
\footnotesize
\begin{tabular}{|c|c|c|c|c|c|c|c|}
\hline
k\_loc = 4 & \multicolumn{7}{c|}{Percent of Edges Failed}    \\  
\hline
Method & 0.00 & 0.01  & 0.02   & 0.05   & 0.10 & 0.20 & 0.25  \\
\hline
\textbf{Baseline Methods} & &  &  &  &  &  & \\ 
Spectral \cite{Doherty_2022}& 0.013(0.030) & 0.084(0.350) & 0.102(0.310) & 0.117(0.477) & 0.270(0.784) & 0.310(0.763) & 0.325(0.833) \\
L1 \cite{Hartley_2011_L1}& 0.014(0.052) & 0.020(0.135) & 0.055(0.502) & 0.080(0.296) & 0.449(\textbf{0.506}) & 0.499(\textbf{0.500}) & 0.490(\textbf{0.501}) \\
\hline
\hline
\textbf{Cauchy Loss(Ours)} & & & & & & & \\
$k=3$ & 0.013(\textbf{0.030}) & 0.013(0.036) & \textbf{0.011}(0.032) & 0.059(0.395) & 0.107(0.507) & 0.217(0.778) & 0.255(0.968) \\
$k=5$ & 0.013(\textbf{0.030}) & 0.013(0.038) & \textbf{0.011}(0.033) & 0.054(0.392) & 0.109(0.513) & 0.206(0.764) & \textbf{0.242}(0.637) \\
$k=8$ & 0.013(\textbf{0.030}) & 0.013(0.037) & \textbf{0.011(0.031)} & \textbf{0.054(0.386)} & 0.107(\textbf{0.506}) &\textbf{ 0.205}(0.640) & 0.267(0.548) \\
\hline
\hline
\textbf{G-MC Loss(Ours)} & & & & & & & \\
$k=3$ & \textbf{0.012(0.030)} & \textbf{0.012(0.034)} & 0.020(0.068) & 0.077(0.371) & 0.100(0.641) & 0.268(0.789) & 0.323(0.723) \\
$k=5$ & \textbf{0.012}(\textbf{0.030}) & \textbf{0.012}(0.035) & 0.017(0.056) & 0.073(0.375) & 0.096(0.642) & 0.235(0.589) & 0.284(0.871) \\
$k=8$ & \textbf{0.012}(\textbf{0.030}) & \textbf{0.012(0.034)} & 0.015(0.056) & 0.064(0.195) & \textbf{0.084}(0.652) & 0.219(0.531) & 0.258(0.599) \\
\hline
\end{tabular}
\end{subtable}
\caption{ \small Comparison of error rates for different methods on varying levels of edge failures of $SO(2)$. The $F$-norm loss $d^{\ell=1}_{F}$ is shown outside of parenthesis, the $d^{\ell=1}_{\infty}$ norm is shown in parenthesis. It is important to acknowledge that within \cite{Hartley_2011_L1}, there is no explicit procedure available for SO(2). The L1 method is predicated on the SO(3) approach outlined in \cite{Hartley_2011_L1}.}
\label{fig:SO2_edge_fail_comparison_II}
\end{figure*}

\begin{figure*}[h!]
\begin{subtable}{0.80\paperwidth}
\footnotesize
\vspace{-1.1cm}
\begin{tabular}{|c|c|c|c|c|c|c|c|}
\hline
k\_loc = 4 & \multicolumn{7}{c|}{Percent of Edges Failed}    \\  
\hline
Method & 0.00 & 0.01  & 0.02   & 0.05   & 0.10 & 0.20 & 0.25  \\
\hline
\textbf{Baseline Methods} & &  &  &  &  &  & \\ 
Spectral \cite{Doherty_2022} & 0.009(0.021) & 0.059(0.248) & 0.072(0.220) & 0.083(0.339) & 0.191(0.562) & 0.220(0.546) & 0.230(0.598) \\
L1\cite{Hartley_2011_L1} & 0.010(0.037) & 0.014(0.095) & 0.039(0.357) & 0.057(0.210) & 0.319(0.360) & 0.355(\textbf{0.355}) & 0.348(\textbf{0.356}) \\
\hline
\hline
\textbf{Cauchy Loss(Ours)} & & & & & & & \\
$k=3$ & 0.009(0.021) & 0.009(0.025) & \textbf{0.008}(0.023) & 0.042(0.280) & 0.076(\textbf{0.360}) & 0.154(0.557) & 0.181(0.699) \\
$k=5$ &  0.009(0.021) & 0.009(0.027) & \textbf{0.008}(0.023) & 0.038(0.278) & 0.077(0.365) & 0.146(0.547) & \textbf{0.171}(0.454) \\
$k=8$ &  0.009(0.021) & 0.009(0.026) & \textbf{0.008(0.022)} & \textbf{0.038(0.274)} & 0.076(\textbf{0.360}) & \textbf{0.145}(0.457) & 0.189(0.390) \\
\hline
\hline
\textbf{G-MC Loss(Ours)} & & & & & & & \\
$k=3$ & \textbf{0.008(0.021)} & \textbf{0.008(0.024)} & 0.014(0.048) & 0.054(0.263) & 0.071(0.457) & 0.190(0.565) & 0.229(0.517) \\
$k=5$ &\textbf{0.008(0.021)} & \textbf{0.008}(0.025) & 0.012(0.040) & 0.052(0.266) & 0.068(0.458) & 0.166(0.420) & 0.190(0.626) \\
$k=8$ & \textbf{0.008(0.021)} & \textbf{0.008(0.024)} & 0.011(0.040) & 0.045(0.138) & \textbf{0.060}(0.465) & 0.155(0.378) & 0.183(0.427) \\
\hline
\end{tabular}
\end{subtable}
\caption{ \small Comparison of angle difference for different methods on varying levels of edge failures of $SO(2)$, expressed in radius. The mean angle difference $\Delta\phi_{mean}$ is shown outside of parenthesis, the maximum angle difference $\Delta\phi_{\max}$ is shown in parenthesis. It is important to acknowledge that within \cite{Hartley_2011_L1}, there is no explicit procedure available for SO(2). The L1 method is predicated on the SO(3) approach outlined in \cite{Hartley_2011_L1}.}
\label{fig:SO2_edge_fail_comparison_III}
\end{figure*}

\begin{figure*}[h!]
\begin{subtable}{0.10\paperwidth}
\footnotesize
\begin{tabular}{|c|c|c|c|c|c|c|c|}
\hline
k\_loc = 8 & \multicolumn{7}{c|}{Percent of Edges Failed}    \\  
\hline
Method & 0.00 & 0.01  & 0.02   & 0.05   & 0.10 & 0.20 & 0.25  \\
\hline
\textbf{Baseline Methods} & &  &  &  &  &  & \\ 
Spectral \cite{Doherty_2022} & \textbf{0.006(0.016)} & 0.035(0.160) & 0.037(0.138) & 0.066(0.22) & 0.106(0.375) & 0.163(0.490) & 0.184(0.547) \\
L1\cite{Hartley_2011_L1} & 0.015(0.024) & 0.015(0.078) & 0.036(0.338) & 0.120(0.490) & 0.083(0.496) & 0.308(0.499) & 0.499(0.499) \\
\hline
\hline
\textbf{Cauchy Loss(Ours)} & & & & & & & \\
$k=3$ & \textbf{0.006}(0.017) & 0.008(0.027) & 0.007(0.019) & 0.015(0.068) & 0.015(0.061) & 0.045(0.205) & 0.096(0.632) \\
$k=5$ & \textbf{0.006}(0.017) & 0.008(0.020) & \textbf{0.006(0.019)} & 0.007(0.030) & 0.009(0.034) & 0.019(0.127) & 0.089(0.719) \\
$k=8$ & \textbf{0.006}(0.017) & 0.008(\textbf{0.019}) & \textbf{0.006(0.019)}& \textbf{0.006(0.020)} & \textbf{0.009(0.029)} & \textbf{0.012(0.046)} & 0.062(\textbf{0.404}) \\
\hline
\hline
\textbf{G-MC Loss(Ours)} & & & & & & & \\
$k=3$ & 0.007(0.018) & \textbf{0.007}(0.023) & 0.011(0.032) & 0.013(0.052) & 0.020(0.082) & 0.061(0.483) & 0.084(0.622) \\
$k=5$ & 0.007(0.018) & \textbf{0.007}(0.022) & 0.008(0.023) & 0.011(0.060) & 0.013(0.060) & 0.051(0.424) & 0.080(0.602) \\
$k=8$ & 0.007(0.018) & \textbf{0.007}(0.022) & 0.008(0.023) & 0.008(0.045) & 0.010(0.034) & 0.026(0.144) & \textbf{0.061}(0.530) \\
\hline
\end{tabular}
\end{subtable}
\caption{ \small Comparison of error rates for different methods on varying levels of edge failures of $SO(2)$. The $F$-norm loss $d^{\ell=1}_{F}$ is shown outside of parenthesis, the $d^{\ell=1}_{\infty}$ norm is shown in parenthesis. It is important to acknowledge that within \cite{Hartley_2011_L1}, there is no explicit procedure available for SO(2). The L1 method is predicated on the SO(3) approach outlined in \cite{Hartley_2011_L1}}
\label{fig:SO2_edge_fail_comparison_IV}
\end{figure*}

\subsection{Numerical Experiments on $SO(3)$ Rotation Averaging}\label{Numerical Experiments on so3 Rotation Averaging}

\begin{figure}[h!]
\centering
\begin{tabular}{|l|}
\hline
\includegraphics[width=0.45\textwidth]{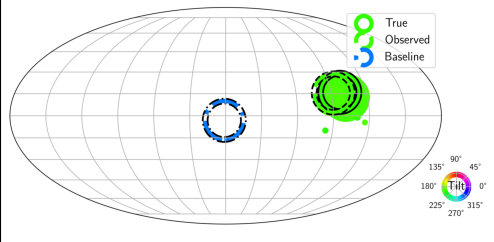}
\includegraphics[width=0.45\textwidth]{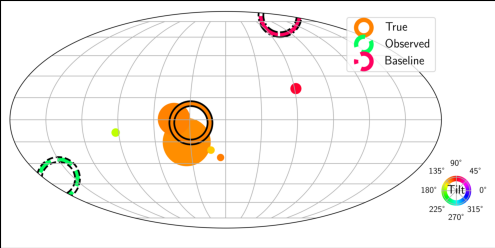} \\
\hline
\end{tabular}
\vspace*{-0.0cm}
\caption{ \small $SO(3)$ Edge Recovery. Left panel shows $|D_{ij}(R)|^{2}$ as a function of rotation $R \in SO(3)$ for an edge that is corrupted by Langevin noise. Left panel shows $|D_{ij}(R)|^{2}$ as a function of rotation $R \in SO(3)$ for an edge that is corrupted by totally random noise. Full circles denote the true difference. Dotted circles denote the solution recovered by \cite{Doherty_2022}. Note that while our method and the \cite{Doherty_2022} method recover similar solutions on the edge corrupted by Langevin noise (left), our method is \emph{also} able to recover accurate estimates for edges corrupted by uniform noise (right). For more information on representing $SO(3)$ rotations on mollweide plots, please see \cite{Murphy_2022_implicitpdf}   }\label{Table:SO3_Predictions}
\end{figure}

We also test our results on three dimensional rotation averaging problems. We specifically test our results on graphs generated by NetworkX library in python. Figure \ref{Small world} shows the nodes and edges of the graphs. Table \ref{fig:SO3_edge_fail_comparison_I} and \ref{fig:SO3_edge_fail_comparison_II} shows the comparison of performance of our method as a function of different number of irreducibles used. Note that in the high noise regime of \ref{fig:SO3_edge_fail_comparison_II} our method outperforms existing baselines by more than a factor of two.

\begin{figure*}[h!]
	\begin{subtable}{0.10\linewidth}
		\small
		\begin{tabular}{|c|c|c|c|c|c|c|}
			\hline
			k\_loc = 5 & \multicolumn{6}{c|}{Percent of Edges Failed} \\  
			\hline
			Method & p=0.05 & p=0.10 & p=0.15 & p=0.20 & p=0.25 & p=0.30 \\
			\hline
			\textbf{Baseline Methods} & &  &  &  &  &  \\ 
			Spectral \cite{Doherty_2022} & 0.078(0.290) & 0.128(0.443) & 0.192(0.454) & 0.206(0.512) & 0.249(0.508) & 0.430(1.064) \\
			L1\cite{Hartley_2011_L1} & 0.036(0.156) & 0.085(0.448) & 0.128(0.473) & 0.143(0.556) & 0.431(0.997) & 0.405(0.998) \\
			Shonan\cite{Dellaert_2020_Shonan} & 0.039(0.221) & 0.127(0.660) & 0.211(0.531) & 0.235(0.612) & 0.481(0.712) & 0.531(0.971)  \\
			\hline
			\hline
			\textbf{Cauchy Loss(Ours)} & & & & & & \\
			$l=3$ & 0.075(0.182) & 0.119(0.340) & 0.112(0.387) & 0.115(0.407) & 0.191(0.418) & 0.288(0.724) \\
			$l=5$ & 0.061(0.115) & 0.093(0.257) & 0.107(0.283) & 0.111(0.402) & 0.191(0.418) & 0.277(0.623) \\
			$l=8$ & 0.063(0.134) & 0.089(0.275) & 0.109(0.333) & 0.116(0.414) & 0.191(0.419) & 0.261(0.643) \\
			\hline
		\end{tabular}
	\end{subtable}
	\caption{ \small Comparison of error rates for different methods on varying levels of edge failures of $SO(3)$. The $F$-norm loss $d^{\ell=1}_{F}$ is shown outside of parenthesis, the $d^{\ell=1}_{\infty}$ norm is shown in parenthesis.}
	\label{fig:SO3_edge_fail_comparison_I}
\end{figure*}

\begin{figure*}[h!]
\begin{subtable}{0.01\linewidth}
\small
\begin{tabular}{|c|c|c|c|c|c|c|}
\hline
k\_loc = 8 & \multicolumn{6}{c|}{Percent of Edges Failed} \\  
\hline
Method & p=0.05 & p=0.10 & p=0.15 & p=0.20 & p=0.25 & p=0.30 \\
\hline
\textbf{Baseline Methods} & &  &  &  &  &  \\ 
Spectral \cite{Doherty_2022} & 0.051(0.147) & 0.068(0.254) & 0.098(0.284) & 0.083(0.274) & 0.140(0.339) & 0.167(0.474) \\
L1\cite{Hartley_2011_L1} & \textbf{0.014(0.040)} & \textbf{0.020}(0.115) &
\textbf{0.034}(0.137) & 0.043(0.246) & 0.073(0.249) & 0.099(0.460) \\
Shonan\cite{Dellaert_2020_Shonan} & 0.040(0.227) & 0.046(0.251) & 
0.068(0.265) & 0.080(0.276) & 0.097(0.340) & 0.120( 0.620 ) \\
Graduated Non-Convexity \cite{Yang_2020_Graduated} & 0.061(0.013) & 0.051(0.191) & 
0.074(0.89) & 0.196(0.92) &  0.337(0.93) & 0.431(0.95) \\ 
\hline
\hline
\textbf{Cauchy Loss(Ours)} & & & & & & \\
$l=3$ & 0.042(0.099) & 0.041(0.101) & 0.044(0.094) & 0.049(0.138) & 0.061(0.129) & 0.084(0.248) \\
$l=5$ & 0.036(0.078) & 0.040(\textbf{0.100}) & 0.037(\textbf{0.088}) & \textbf{0.042(0.112)} & \textbf{0.040(0.097)} & \textbf{0.045(0.160)} \\
$l=8$ & 0.036(0.080) & 0.036(0.085) & 0.041(0.090) & 0.045(0.125) & 0.053(0.102) & 0.050(0.172) \\
\hline
\end{tabular}
\end{subtable}
\caption{Comparison of error rates for different methods on varying levels of edge failures of $SO(3)$. The $F$-norm loss $d^{\ell=1}_{F}$ is shown outside of parenthesis, the $d^{\ell=1}_{\infty}$ norm is shown in parenthesis. For fair comparison Shonan \cite{Dellaert_2020_Shonan} and Graduated Non-Convexity \cite{Yang_2020_Graduated} were initialized using the initialization of \cite{Hartley_2011_L1}. Each result is averaged over $M=10$ trials.   }
\label{fig:SO3_edge_fail_comparison_II}
\end{figure*}

\end{document}